\documentclass{article}

\PassOptionsToPackage{numbers}{natbib}
\usepackage[preprint]{neurips_2026}
\workshoptitle{AI for Drug Discovery -- Bridging the Translation Gap}

\usepackage[utf8]{inputenc}
\usepackage[T1]{fontenc}
\usepackage{hyperref}
\usepackage{url}
\usepackage{booktabs}
\usepackage{amsfonts,amsmath,amssymb,amsthm}
\usepackage{nicefrac}
\usepackage{microtype}
\usepackage{xcolor}
\usepackage{bm}
\usepackage{graphicx}
\usepackage{tikz}
\usepackage{pifont}
\usepackage{placeins}
\usepackage{etoc}
\newcommand{\xmark}{\ding{55}}
\usetikzlibrary{arrows,arrows.meta,shapes,positioning,calc}

\definecolor{cA}{HTML}{1D9E75}
\definecolor{cB}{HTML}{7F77DD}
\definecolor{cone}{HTML}{C6803A}
\definecolor{seg}{HTML}{888780}
\definecolor{cstar}{HTML}{2B2A27}
\definecolor{xpt}{HTML}{D85A30}
\colorlet{myblue}{cA!70}
\colorlet{mybluelight}{cA!42}
\colorlet{myred}{xpt!80}
\colorlet{ctA}{cA}\colorlet{ctB}{cB}\definecolor{navy}{HTML}{13294B}\definecolor{muted}{HTML}{55657E}

\newtheorem{proposition}{Proposition}

\title{Cell-Type-Specific Expression Is Hard to Predict from Bulk, but It Is Still Useful}

\author{
  Gadi Naveh\thanks{Corresponding author.} \qquad Amit Miller \qquad Gal Keinan \\
  GSK.ai \\
  144 Menahem Begin Rd., Tel Aviv, Israel \\
  \texttt{\{gadi.x.naveh, amit.x.miller, gal.x.keinan\}@gsk.com}
}

\begin{document}

\maketitle

\begin{abstract}
  Single-cell-informed deconvolution provides cell-type-resolved features for clinical bulk RNA cohorts.
  Most existing methods predict the \textit{proportions} $\pi_k$ of the $K$ constituent cell types (CTs). 
  Fewer predict \textit{cell-type-specific expression} (CTSE) $\mu_{kg}$ of gene $g$ in CT $k$, which can capture cell states that proportions miss, but recent benchmarks find it much harder.
  We explain this gap with closed-form accuracy bounds.
  For CTSE, we derive an $R^2$ ceiling for per-gene linear estimators.
  For proportions, the best linear estimator, which pools evidence across genes, has a closed-form $R^2$, and a computable lower bound on it predicts which CTs are learnable.
  We study \textit{targeted deconvolution} of 106 biologically curated (CT, gene) pairs, found post hoc to have higher learnability scores.
  On overall survival in non-small cell lung cancer, estimation difficulty does not predict prognostic value across these targets.
  Yet a sparse survival model on deconvolved features, mostly CTSE, predicts survival as well as a sparse model on the bulk genes themselves, while attributing its signal to specific cell types. Some deconvolved CTSE features reveal prognostic signal not seen in the corresponding bulk gene.
  \end{abstract}

\etocsettocdepth.toc{none}
\section{Introduction}

\begin{figure}[!t]
\centering
\includegraphics[width=\textwidth]{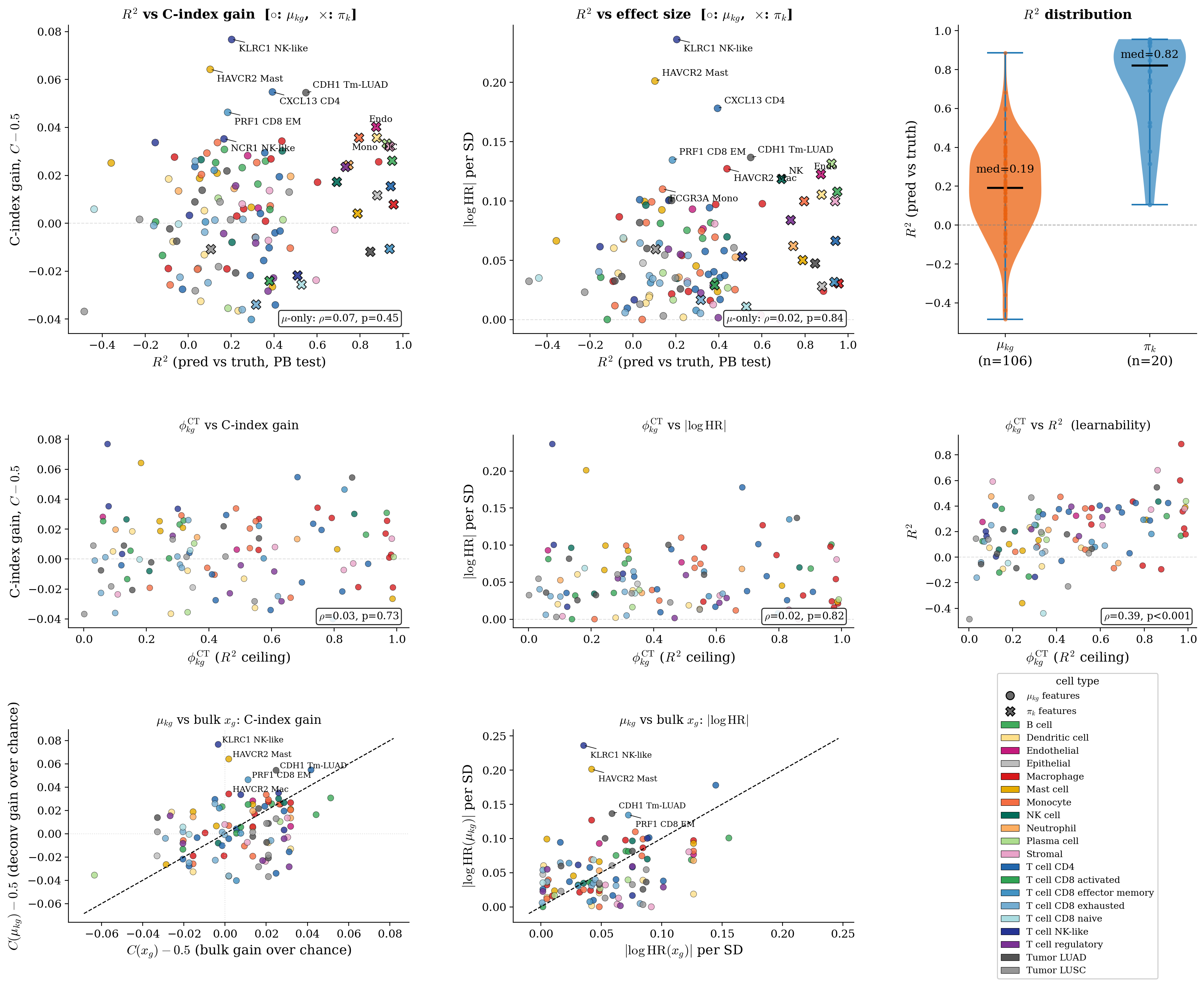}
\caption{\textbf{$\phi^{\mathrm{CT}}_{kg}$ tracks accuracy, but not prognostic value.}
Points are predicted CTSE ($n{=}106$; $\circ$) or CT proportions ($n{=}20$; $\times$), coloured by cell type.
\textbf{Top row.}
\emph{Left/centre}: $R^2$ of deconvolution predictions on pseudobulk test data plotted against cross-validated C-index gain ($C_{\rm deconv}{-}0.5$) and effect size ($|\log\mathrm{HR}|$ per SD of the feature; univariate Cox) on TCGA NSCLC OS.
Both are uncorrelated with CTSE accuracy (Spearman $\rho{=}0.07$, $p{=}0.45$ and $\rho{=}0.02$, $p{=}0.84$).
\emph{Right}: median $R^2$ is $0.19$ for log-transformed CTSE vs $0.82$ for proportions.
\textbf{Middle row.}
The linear-CPM score $\phi^{\mathrm{CT}}_{kg}$ correlates with our MLP's CTSE accuracy ($\rho{=}0.39$, $p{<}0.001$), but is uncorrelated with C-index gain ($\rho{=}0.03$, $p{=}0.73$) and $|\log\mathrm{HR}|$ ($\rho{=}0.02$, $p{=}0.82$).
\textbf{Bottom row.}
Paired comparison of predicted CTSE against bulk expression of the \emph{same gene}: gain over chance ($C{-}0.5$; left) and effect size (right), for all pairs with a bulk counterpart. On average the two are comparable (median $\Delta C\approx0$); some CTSE features, e.g.\ KLRC1 in NK-like T cells, carry signal not apparent in the bulk gene.
Labelled examples are post hoc picks from our primary model (robustness: App.~\ref{app:features}).}
\label{fig:combined}
\end{figure}

Bulk deconvolution decomposes a gene expression vector $\bm{x}^{(i)} \in \mathbb{R}^G$ as
\begin{equation}
  \label{eq:exact_lin}
  \bm{x}^{(i)} = \sum_{k=1}^{K}\pi_k^{(i)}\,\bm{\mu}_k^{(i)}
\end{equation}
where $\pi_k^{(i)}$ is CT $k$'s fraction of measured RNA and $\bm{\mu}_k^{(i)} \in \mathbb{R}^G$ is its mean expression profile (CTSE) in sample $i$. Its variation across samples captures within-CT variation (e.g.\ activation or exhaustion), which proportions do not.
See App.~\ref{app:notation} for precise conventions on proportion normalization ($\pi^{\mathrm{lib}}$ vs $\pi^{\mathrm{count}}$) and expression space (linear vs log).
Most methods predict $\pi_k$, and typically do so well. In the DREAM challenge the best methods reach CT-averaged Pearson correlations of $0.85{-}0.90$ at coarse CT resolution and $0.76$ at fine resolution~\cite{white2024dream}; on pseudobulk, errors are small (median RMSE ${<}0.05$)~\cite{avilacobos2020benchmarking}. In contrast, a recent CTSE benchmark reports a best mean Spearman correlation of only $0.4$, comparable to a naive bulk-only baseline~\cite{hu2025daleeval}. Performance improves on CT-specific marker genes, which, as we show below, contribute a larger share of bulk variation at their gene.

Why is CTSE harder to recover? Intuitively, each bulk gene is a single mixture of $K$ CT-specific values, whereas proportions can pool evidence across the many genes that distinguish the CTs. For estimators using one gene and known composition, we derive an accuracy ceiling $\phi^{\mathrm{CT}}_{kg}$ (\S\ref{sec:ceiling}). Under independent CT variation, it reduces to the \textit{variance share} $\phi_{kg}$: the fraction of residual bulk variance at gene $g$ contributed by CT $k$. Five of the seven methods benchmarked in~\cite{hu2025daleeval} use per-gene linear estimators, so the ceiling applies to their oracle versions (App.~\ref{app:dale_methods}). Whole-profile models can use additional cross-gene information; for them, these quantities serve as measures of difficulty rather than ceilings.
Because CTSE captures cell states that proportions miss, it may also carry clinical information~\cite{topalian2012}; we test whether hard-to-learn targets still predict NSCLC survival.

\textbf{Contributions.} 
(1) \emph{Why CTSE is hard:} a closed-form ceiling $\phi^{\mathrm{CT}}_{kg}$ on the $R^2$ of per-gene linear CTSE estimators, reducing to the variance share $\phi_{kg}$ under independent CT variation (Prop.~\ref{prop:ceiling_intro}): attained by the posterior mean~\cite{rahmani2019, chen2025unico}, information-theoretic under Gaussianity, and computable from an atlas, where it tracks learnability.
(2) \emph{Why proportions are easier:} the best linear proportion estimator has a closed-form $R^2$, $\eta_k$, pooling evidence across genes (Prop.~\ref{prop:varrho}); its computable lower bound tracks which CTs are learnable, giving a principled way to set CT resolution.
(3) \emph{Targeted deconvolution:} 106 biologically curated (CT, gene) pairs, mainly immune checkpoints and tumour-microenvironment markers, found post hoc to have higher learnability scores.
(4) \emph{Why CTSE is still useful:} on NSCLC survival, estimation difficulty does not predict prognostic value. A sparse model of interpretable deconvolved features matches bulk-gene LASSO on the same inputs; some CTSE features carry signal not apparent in the matched bulk gene (e.g., KLRC1 in NK-like T cells).

\subsection{Related Work}

Benchmarks for predicting $\pi$ ~\cite{white2024dream} and $\mu$~\cite{hu2025daleeval, dai2024sampledeconv} focus mostly on average accuracy, not target selection or downstream utility. 
Existing CTSE methods include per-gene linear estimators given proportions and fitted parameters (BayesPrism~\cite{chu2022bayesprism}, InstaPrism~\cite{hu2024instaprism}, TCA~\cite{rahmani2019}, Unico~\cite{chen2025unico}, bMIND~\cite{wang2021bmind}, EPIC-unmix~\cite{tang2025epicunmix}), per-gene nonlinear estimators (CIBERSORTx~\cite{newman2019cibersortx}, VarNMF~\cite{fallik2025varnmf}) and gene-coupled methods (ENIGMA~\cite{wang2023enigma}); Table~\ref{tab:dale_methods} gives their estimators and scope. Deep CTSE methods include TAPE~\cite{chen2022tape}, DISSECT~\cite{khatri2024dissect}, BLUE~\cite{zhu2026blue} and DESCENT~\cite{zhao2026descent}; DeSide~\cite{xiong2024deside} and DECODE~\cite{zhao2026decode} predict $\pi$ only.
TCA~\cite{rahmani2019} derives the conditional-mean estimator and notes its reduced posterior variance in an equal-variance special case (its Suppl.\ Eq.~16); Unico~\cite{chen2025unico} generalises the estimator to full cross-CT covariance. $\phi^{\mathrm{CT}}_{kg}$ expresses this as the population $R^2$ of that estimator, which we use as a per-target ceiling. Concurrent work ranks CT--gene pairs by a learned recoverability score~\cite{kang2026cdseq2}; $\phi^{\mathrm{CT}}_{kg}$ is instead analytic and atlas-derived. For proportions, Fisher-information and covariance-aware uncertainty have been derived~\cite{erdmannpham2021rnasieve, cai2024decals}; Prop.~\ref{prop:varrho} gives the population $R^2$ of the best linear predictor. 
DESCENT~\cite{zhao2026descent} first generates a cellular representation from bulk and then combines it with bulk for survival prediction; here, CTSE predictions are evaluated separately from outcome-supervised training.
Purified tumour expression (ISOpure~\cite{quon2013isopure}) and CTSE of selected genes (PRISM~\cite{hakkinen2021prism}) have outperformed bulk expression for prognosis, while EcoTyper~\cite{luca2021ecotyper} and Statescope~\cite{janssen2026statescope} relate expression-defined cell-state abundances to survival. We compare predicted CTSE with matched bulk genes and CT proportions, assessing learnability and downstream utility separately.

\section{Methods and Results}

\subsection{Closed-Form Bounds for Expression and Proportion Estimation}
\label{sec:ceiling}

How accurately can $\mu_{kg}$ be recovered from a single bulk gene when the composition is known?

\begin{proposition}[Expression upper bound for per-gene estimators]
\label{prop:ceiling_intro}
Assume no assay noise, and that the deviations $\delta_{kg}^{(i)} := \mu_{kg}^{(i)} - \mu_{kg}^*$ are independent of composition and jointly Gaussian with covariance $\left[\bm{\Sigma}_{g}\right]_{kk'}=\operatorname{Cov}\left(\delta_{kg},\delta_{k'g}\right)$.
Assume oracle knowledge of the CT centroids $\mu_{kg}^*$ and proportions $\pi_k$, and define the bulk residual $e_g^{(i)} := x_g^{(i)} - \sum_k \pi_k^{(i)} \mu_{kg}^* = \sum_k \pi_k^{(i)} \delta_{kg}^{(i)}$.
Then any estimator $\hat{\mu}_{kg}^{(i)}$ conditioning only on $(e_g^{(i)}, \bm{\pi}^{(i)})$ satisfies, for all $k{=}1,\ldots,K$ simultaneously,
\begin{equation}
\label{eq:phi_ct}
    R^2(\hat{\mu}_{kg}) \le \mathrm{corr}^2(\hat{\mu}_{kg}^{(i)}, \mu_{kg}^{(i)}) \le \phi^{\mathrm{CT}}_{kg}, \qquad
    \phi^{\mathrm{CT}}_{kg} := \frac{[(\bm{\Sigma}_g\bm{\pi})_k]^2}{[\bm{\Sigma}_g]_{kk} \cdot (\bm{\pi}^{\!\top}\bm{\Sigma}_g\bm{\pi})}.
\end{equation}
For independent cell types, $\bm{\Sigma}_g = \operatorname{diag}(\tau_{1g}^2,\ldots,\tau_{Kg}^2)$, this reduces to the \emph{variance share}
\begin{equation}
\label{eq:phi_intro}
    \phi_{kg} := \frac{\pi_k^2 \tau_{kg}^2}{\sum_{k'} \pi_{k'}^2 \tau_{k'g}^2}, \qquad \sum_k \phi_{kg} = 1.
\end{equation}
All moments and $R^2$ here condition on the sample's composition $\bm{\pi}$, and $\tau_{kg}^2 = \operatorname{Var}(\mu_{kg})$. Pooled performance across compositions is discussed in App.~\ref{app:proof_ceiling}.
\end{proposition}

\textbf{Tightness.} The Gaussian conditional mean $\mathbb{E}[\mu_{kg}|e_g,\bm{\pi}]$ attains the ceiling for all $K$ CTs simultaneously. Under Gaussianity this is an information-theoretic limit for estimators using only the gene and composition (App.~\ref{app:proof_ceiling}).

\textbf{Scope.} Without Gaussianity, the ceilings remain exact for estimators linear in $e_g$ at fixed composition (App.~\ref{app:varrho:floor}). A gradient-boosted predictor using only the gene and the composition improves little on the atlas-based Gaussian-form predictor (median $+0.007$ held-out $R^2$), although some targets benefit substantially. Whole-profile models, including our MLP, are outside this class, so the bound does not apply to them. At uniform composition and equal variances, $\phi_{kg}=1/K$: an $R^2$ ceiling of $0.05$ at our $K{=}20$, and $0.13{-}0.2$ at the coarser resolutions ($K{\approx}5{-}8$) benchmarked in DALE-Eval~\cite{hu2025daleeval}, where accuracy is also low.
Figures use scores evaluated at mean composition; these plug-in values differ from the pooled ceiling across compositions (App.~\ref{app:proof_ceiling}).

\textbf{Why proportions are easier.} Per-gene variance shares are mostly small (median $\phi_{kg}=0.07$ over our pairs), but proportions can pool evidence across genes. With two CTs, Eq.~\eqref{eq:exact_lin} reads $\bm{x}=\bm{\mu}_2^*+\pi\bm{\Delta}+\bm{e}$, where $\pi:=\pi_1$, $\bm{\Delta}:=\bm{\mu}_1^*-\bm{\mu}_2^*$ is the centroid contrast and $\bm{e}:=\sum_k\pi_k(\bm{\mu}_k-\bm{\mu}_k^*)$ is the bulk residual (Fig.~\ref{fig:tau_phi}). Its covariance $\bm{T}_e:=\mathrm{Cov}(\bm{e})$ averages over the composition distribution, so it does not depend on the predicted sample (App.~\ref{app:varrho:main}).
\begin{proposition}[Proportion $R^2$ for $K{=}2$]
\label{prop:varrho}
If $\mathrm{Cov}(\bm{e},\pi)=\bm{0}$, $v>0$ and $\bm{T}_e$ is positive definite, any predictor $\hat\pi$ linear in $\bm{x}$ satisfies, with equality for the best one,
\begin{equation}
\label{eq:r2_pi}
  R^2(\hat\pi) \le \eta := \frac{v\,\varrho^2}{1+v\,\varrho^2}, \qquad
  \varrho^2 := \bm{\Delta}^{\!\top}\bm{T}_e^{-1}\bm{\Delta}, \qquad v := \mathrm{Var}(\pi).
\end{equation}
For $K$ CTs, the per-CT proportion $R^2$, $\eta_k$, has a $(K{-}1)$-dimensional analogue (Eq.~\eqref{eq:pi_err_cov}, App.~\ref{app:varrho}).
\end{proposition}
Here $\varrho$ measures centroid separation relative to residual variation. It pools evidence across genes that distinguish the CTs, unlike the per-gene expression ceiling. $\eta$ is the best achievable linear $R^2$. Because $(\pi,\bm{x})$ cannot be jointly Gaussian ($\pi$ is bounded and the residual variance depends on $\pi$), $\eta$ need not bound nonlinear methods: an optimal nonlinear predictor can match or exceed it. Closely related CTs or large expression variation along their contrasts can make proportions difficult too.

\subsection{Experimental Methods}

\textbf{Model.} A skip-MLP trunk (3 layers of 512 units) maps $G{=}1024$ log-transformed input genes to a shared representation $\bm{h}$. Proportion heads predict $\hat{\bm{\pi}}$ from $\bm{h}$ (sigmoid followed by simplex renormalisation; $K{=}20$ CTs). Each CTSE head combines $\bm{h}$ with the matched gene's log-bulk expression through a skip connection. Training uses KL divergence for proportions and masked MSE for log-CTSE, with auxiliary heads detailed in App.~\ref{app:methods}.

\textbf{Training data.} 
Pseudobulk is generated from an NSCLC single-cell atlas~\cite{salcher2022nsclc} (${\sim}1.28$M cells; 512 samples with ${\ge}100$ cells, from 307 donors). For each atlas sample we draw about 100 compositions from a Dirichlet distribution centred on the sample's own composition over 51 fine-grained CTs (concentration $\alpha{=}30$), resample that sample's cells with replacement to match each composition and its total cell count, and sum raw counts before CPM normalisation (App.~\ref{app:notation}). Targets are the $K{=}20$ proportions (aggregating the 51 fine CTs) and 106 $\mu_{kg}$; train and test sets are split by atlas sample; a donor-held-out replicate gives the same results (App.~\ref{app:robust}).

\textbf{Downstream analysis.} Models applied to TCGA NSCLC ($n{=}998$, 394 events). Univariate Cox per feature (HR per SD), with Benjamini--Hochberg correction within each family; cross-validated LASSO-Cox on ensemble features for combinations. Model-robustness and adjustment checks are in App.~\ref{app:controls}; methods in App.~\ref{app:methods}.

\subsection{Results}
\label{sec:results}

\textbf{Estimation accuracy.} Median pseudobulk test $R^2$ is $0.82$ for proportions and $0.19$ for log-transformed CTSE (a per-target ridge on the same inputs reaches $0.28$ on donor-held-out samples; App.~\ref{app:robust:c2}); we report $R^2$ because, unlike correlation, it penalizes bias and scale errors (App.~\ref{app:notation:r2}). After back-transformation to linear CPM, where the ceilings are defined, squared correlations also track $\phi_{kg}$ and $\phi^{\mathrm{CT}}_{kg}$ ($\rho{=}0.53$ and $0.52$; App.~\ref{app:notation:r2}).

\textbf{Learnability and observed accuracy.} Both atlas-derived scores correlate with the MLP's per-target test $R^2$ (Spearman $\rho{=}0.49$ for $\phi_{kg}$ and $0.39$ for $\phi^{\mathrm{CT}}_{kg}$, $p{<}0.001$; Fig.~\ref{fig:combined}), and with that of the per-target ridge ($0.65$ and $0.51$; App.~\ref{app:robust:c2}). Per-gene linear estimators, the class Prop.~\ref{prop:ceiling_intro} covers, stay below $\phi^{\mathrm{CT}}_{kg}$ for 92--97 of 106 targets and track it ($\rho{\approx}0.6$; App.~\ref{app:robust}). For proportions, the projected lower bound $\eta_k^{W}$ on optimal linear accuracy (App.~\ref{app:varrho:eta_w}) ranks CTs by observed accuracy ($\rho{=}0.76$, $p{=}2{\times}10^{-4}$, 19 evaluable CTs). The least separated pair is CD8 naive vs CD8 effector memory T cells (App.~\ref{app:varrho:merge}). Ridge regression on linear-CPM inputs, the linear class of Prop.~\ref{prop:varrho}, tracks $\eta_k^{W}$ closely ($\rho{=}0.81$) and is on par with our model (median $R^2$ $0.85$ vs $0.82$); the model is better for NK and NK-like T cells, ridge for others (e.g.\ neutrophils, mast cells).

\begin{figure}[h]
\centering
\includegraphics[width=0.95\textwidth]{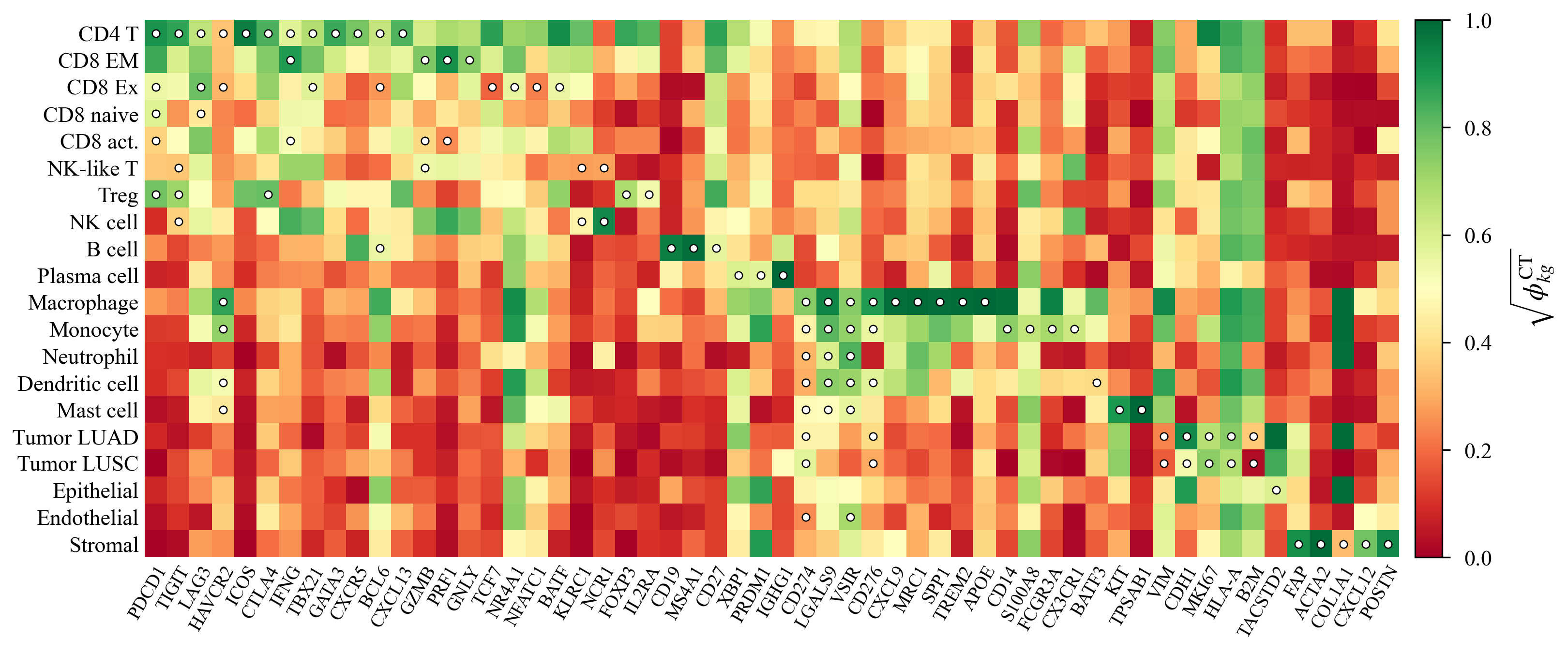}
\caption{\textbf{Atlas-derived per-gene learnability across 20 cell types and 56 target genes.} Color shows $\sqrt{\phi^{\mathrm{CT}}_{kg}}$, incorporating same-gene cross-CT covariance at mean composition. White dots mark the 106 selected pairs. Abbreviations and the independent-CT baseline: Fig.~\ref{fig:phi_diag}.
}
\label{fig:phi_ct}
\vspace{6pt}
\end{figure}

\textbf{Difficulty vs.\ clinical value.} Across the selected targets, neither $\phi^{\mathrm{CT}}_{kg}$ nor pseudobulk accuracy is correlated with prognostic value ($|\log\mathrm{HR}|$ per SD and C-index gain; $|\rho|\le0.07$, Fig.~\ref{fig:combined}): learnability and downstream utility need separate assessment.
Most individual associations are weak after multiplicity correction. KLRC1 in NK-like T cells is the strongest CTSE example meeting our model-robustness criterion (App.~\ref{app:controls}): ensemble HR $0.84$ per SD (95\% CI $0.76{-}0.93$), versus $0.97$ for bulk KLRC1; its association survives BH correction after adjusting for $\hat\pi_k$, $x_g$ and proliferation (HR $0.76$, $q{=}5{\times}10^{-4}$) and remains after further adjustment for age, sex, stage and histology (HR $0.79$, $0.69{-}0.89$). KLRC1 encodes NKG2A, an inhibitory checkpoint targeted clinically~\cite{andre2018nkg2a} and expressed in lung tumours mainly by CD8$^+$ T cells~\cite{chen2020nkg2a}, illustrating how targeted deconvolution can generate cell-type-specific hypotheses about drug-target expression in bulk cohorts. Other CTSE features meeting the (nominal) criterion include PRF1 in CD8 effector memory and IL2RA in regulatory T cells; among proportions, Endothelial is the most consistent (HR $1.14$).
Ensemble LASSO-Cox selects ${\sim}15$ deconvolved features, mostly CTSE, and reaches $C{=}0.58$, similar to bulk-gene LASSO on the same 1{,}024 input genes ($0.57$; difference $+0.014$, 95\% CI $-0.010$ to $0.040$) and above the 55 target genes' bulk expression plus proportions ($+0.031$, $0.008{-}0.052$; both sparse, App.~\ref{app:controls}).

\section{Discussion}

\textbf{Rules of thumb.} 
Our results suggest three practical rules for targeted deconvolution.
(i) Before training, compute $\phi_{kg}$ and $\phi^{\mathrm{CT}}_{kg}$ from an atlas to gauge how well a drug target's CT-specific expression can be read from bulk cohorts, avoiding values near 0 (unlearnable) or 1 (CTSE $\approx$ bulk);
$\phi^{\mathrm{CT}}_{kg}$ caps per-gene linear CTSE methods, and both rank targets by difficulty for richer models; with few reference donors, the target CT's share of mean expression is a usable proxy (App.~\ref{app:robust}). 
(ii) Set CT resolution with $\eta_k$; low-SNR contrasts show which CTs to merge (App.~\ref{app:varrho:merge}).
(iii) Assess downstream utility separately from estimation accuracy: hard-to-predict targets can still carry prognostic signal.

\textbf{Distribution shift and estimated inputs.} 
Both bounds hold within the training distribution and assume known centroids and covariances (and, for CTSE, known composition). Atlas-to-bulk shift (dissociation bias, disease stage, sequencing platform and protocol) can degrade accuracy, and pseudobulk performance may overstate transfer to real bulk~\cite{hippen2023ovarian, dietrich2026omnideconv}. Composition errors also propagate into CTSE predictions (App.~\ref{app:error_propagation}); some methods model this uncertainty explicitly~\cite{cheng2026ubd}.

\textbf{Limitations and outlook.} 
Deconvolved features are functions of bulk, so they add no outcome information beyond that profile; the single-cell-trained map offers a CT-resolved representation for interpretation and downstream modelling. Survival discrimination is modest, and the associations are exploratory rather than clinically validated. Because random bulk-derived signatures can also be prognostic~\cite{venet2011random}, we adjust for abundance, bulk expression and proliferation and require robustness across models; attributing the remaining signal to CT-specific biology requires independent cell-resolved or orthogonal validation, such as matched bulk and single-cell or spatial data~\cite{dai2024sampledeconv, lin2025penalised}. Next steps include independent validation and comparisons with established CTSE methods~\cite{chu2022bayesprism, chen2025unico} and cell-state fractions~\cite{luca2021ecotyper}. Theory extensions include ceilings for models that pool a few correlated genes (App.~\ref{app:cross_ct}), and pseudobulk generators that preserve composition--state coupling (App.~\ref{app:varrho:covary}).

\section*{Acknowledgments}
We thank Alon Brutzkus and Nick Wreglesworth for valuable discussions and Martino Mansoldo for the initial coding infrastructure.

\newpage
\bibliographystyle{plainnat}
\bibliography{references_submission}  %

\newpage
\appendix

\etocsettocdepth.toc{subsection}
\begingroup
\etocsettocstyle{\noindent\textbf{Appendix contents}\par\smallskip\small}{\par}
\tableofcontents
\endgroup
\bigskip

\newpage
\section{Notation and Conventions}
\label{app:notation}

\subsection{Cell-count versus library-weighted proportions}
\label{app:notation:pi}

Two conventions for aggregating single-cell data into pseudobulk give \emph{algebraically exact} but different mixture identities. Let $\mathcal{C}$ be the cells in a pseudobulk sample, $\mathcal{C}_k \subseteq \mathcal{C}$ those of cell type $k$, $r_{cg}$ the raw UMI count of gene $g$ in cell $c$, and $L_c = \sum_{g'} r_{cg'}$ its library size.

\paragraph{Library-weighted (sum-first, used throughout).}
Aggregate raw counts first, then CPM-normalize:
\begin{equation}
\label{eq:defs_lib}
x_g = \frac{\sum_{c \in \mathcal{C}} r_{cg}}{\sum_{c \in \mathcal{C}} L_c} \times 10^6, \quad
\mu_{kg} = \frac{\sum_{c \in \mathcal{C}_k} r_{cg}}{\sum_{c \in \mathcal{C}_k} L_c} \times 10^6, \quad
\pi_k^{\mathrm{lib}} = \frac{\sum_{c \in \mathcal{C}_k} L_c}{\sum_{c \in \mathcal{C}} L_c}.
\end{equation}
Each cell is weighted by its library size $L_c$, so composition measures mRNA fractions. The mixture identity $\sum_k \pi_k^{\mathrm{lib}} \mu_{kg} = x_g$ holds exactly for every gene $g$ and sample. Because bulk RNA-seq also measures mRNA fractions (total mRNA per sample, normalized), this convention approximates the bulk setting, subject to type-specific capture and protocol differences.

\paragraph{Cell-count (CPM-first).}
CPM-normalize per cell, then average. Writing $u_{cg} = 10^6 r_{cg}/L_c$:
\begin{equation}
\label{eq:defs_count}
\tilde{x}_g = \frac{1}{|\mathcal{C}|}\sum_{c \in \mathcal{C}} u_{cg}, \quad
\tilde{\mu}_{kg} = \frac{1}{|\mathcal{C}_k|}\sum_{c \in \mathcal{C}_k} u_{cg}, \quad
\pi_k^{\mathrm{count}} = \frac{|\mathcal{C}_k|}{|\mathcal{C}|}.
\end{equation}
Each cell receives equal weight regardless of library size. The mixture identity $\sum_k \pi_k^{\mathrm{count}} \tilde\mu_{kg} = \tilde x_g$ is also exact. The two conventions are related by per-CT mean library sizes $\ell_k$:
\begin{equation}
\label{eq:link}
\pi_k^{\mathrm{lib}} = \pi_k^{\mathrm{count}}\,\frac{\ell_k}{\bar\ell}, \qquad
\mu_{kg} - \tilde\mu_{kg} = \frac{\mathrm{Cov}_{c \in \mathcal{C}_k}(L_c,\, u_{cg})}{\ell_k},
\end{equation}
where $\bar\ell = \sum_k \pi_k^{\mathrm{count}} \ell_k$. In our atlas, $\ell_k/\bar\ell$ spans roughly $18\times$ across cell types, so mixing conventions (e.g., $\pi^{\mathrm{lib}}$ with $\tilde\mu_{kg}$) produces residuals larger than the signal of interest.

\paragraph{Convention.}
Throughout we use the library-weighted convention and drop the superscript: $\pi_k := \pi_k^{\mathrm{lib}}$, $\mu_{kg}$ as in Eq.~\eqref{eq:defs_lib}. The DALE-Eval benchmark~\cite{hu2025daleeval} uses the cell-count convention ($\tilde\mu_{kg}$, $\pi_k^{\mathrm{count}}$); the variance-share ceiling (Prop.~\ref{prop:ceiling_intro}) holds under both conventions but numerical values differ due to Eq.~\eqref{eq:link}.

\subsection{Linear versus log space}
\label{app:notation:log}

The mixture identity $x_g = \sum_k \pi_k \mu_{kg}$ is exact in \emph{linear} CPM space but not in log space. Training operates on $\log_2(\cdot+1)$-transformed targets ($x_g^{\log} = \log_2(x_g+1)$, $\mu_{kg}^{\log} = \log_2(\mu_{kg}+1)$) for numerical stability. Since $\log_2(x+1)$ is concave, Jensen's inequality gives:
\begin{equation}
\label{eq:jensen}
\underbrace{\log_2\!\left(\sum_k \pi_k \mu_{kg}^{(i)} + 1\right)}_{x_g^{\log,(i)}}
\;\geq\;
\underbrace{\sum_k \pi_k \log_2\!\left(\mu_{kg}^{(i)} + 1\right)}_{\text{naive log-space mixture}}.
\end{equation}
The gap is not a small correction: for CT-specific markers (one CT at $M$ CPM, others silent) with $\pi_k \approx 0.05$--$0.15$ and $M \approx 500$--$1000$ CPM, it is $4$--$6\ \log_2$ units. The mixture identity is exact for aggregated observed counts, but finite-cell sampling and assay variation remain part of the observed expression variability. Both inputs and CTSE targets are log-transformed for training; the model's reported CTSE $R^2$ is in this log space. The ceiling is derived for linear CPM expression, so its association with log-space MLP accuracy is an empirical comparison, not a direct test of the ceiling. The per-gene predictor comparison in App.~\ref{app:varrho:floor} instead uses linear CPM targets.

\subsection{Correlation versus $R^2$}
\label{app:notation:r2}

We report $R^2 = 1 - \mathrm{MSE}/\mathrm{Var}$ of the estimates against the truth. The squared Pearson correlation equals the $R^2$ after the best affine recalibration $a + b\,\hat\mu$; bias or scale error adds to the MSE, so $R^2 \le \mathrm{corr}^2$ for each target. This does not justify squaring a benchmark's mean correlation to obtain its mean $R^2$. We therefore retain benchmark correlations as reported. Model accuracy uses log-transformed CTSE and library-weighted proportions; theoretical expression ceilings and the per-gene comparison use linear CPM CTSE. On donor-held-out test samples, a per-target affine calibration fitted on validation donors raises median linear-CPM $R^2$ from $0.006$ to $0.139$ (App.~\ref{app:robust}). Back-transformed predictions have median linear-CPM $R^2=-0.012$, versus median squared correlation $0.232$: the latter describes an ideal affine recalibration using test-target moments, not an independently validated calibrated predictor, and log-space training need not yield calibrated linear-space estimates.

\section{Expression Estimation: Proofs and Details}
\label{app:geometry}

\subsection{Proof of Proposition~\ref{prop:ceiling_intro}}
\label{app:proof_ceiling}

We prove the ceiling for per-gene estimators conditioning on $(e_g,\bm{\pi})$, first for diagonal covariance, then for general $\bm{\Sigma}_g$. Unless stated otherwise, composition is fixed and all moments and performance measures are conditional on it. Independence of deviations and composition makes their mean and covariance invariant across compositions. Ratios below presume positive target and residual variances; $R^2$ is undefined for constant targets.

\textbf{Setup.}
Define the deviations $\delta_{kg}^{(i)} = \mu_{kg}^{(i)} - \mu_{kg}^*$ where $\mu_{kg}^*$ is the population mean. The bulk residual
\begin{equation}
  e_g^{(i)} = x_g^{(i)} - \sum_k \pi_k^{(i)} \mu_{kg}^* = \sum_k \pi_k^{(i)} \delta_{kg}^{(i)}
\end{equation}
provides one scalar observation constraining $K$ unknowns $\delta_{kg}^{(i)}$.

\textbf{Diagonal case} ($\bm{\Sigma}_g = \operatorname{diag}(\tau_{1g}^2,\ldots,\tau_{Kg}^2)$):

\textbf{Step 1.} Under independence: $\operatorname{Cov}(\delta_{kg}, e_g) = \pi_k \tau_{kg}^2$, $\operatorname{Var}(e_g) = \sum_{k'} \pi_{k'}^2 \tau_{k'g}^2$.

\textbf{Step 2.} For jointly Gaussian $(X,Y)$,
\begin{equation}
\label{eq:gauss_cond}
  \mathbb{E}[X|Y] = \mathbb{E}[X] + \frac{\operatorname{Cov}(X,Y)}{\operatorname{Var}(Y)}(Y - \mathbb{E}[Y])
\end{equation}
Since $\mathbb{E}[\delta_{kg}] = \mathbb{E}[e_g] = 0$:
\begin{equation}
\label{eq:mmse}
\hat{\delta}_{kg}^{(i)} = \mathbb{E}[\delta_{kg} | e_g] = \frac{\pi_k \tau_{kg}^2}{\sum_{k'} \pi_{k'}^2 \tau_{k'g}^2} e_g = \frac{\phi_{kg}}{\pi_k} e_g.
\end{equation}

\textbf{Step 3.} $R^2$ via variance decomposition:
\begin{equation}
\label{eq:r2}
\operatorname{Var}(\mathbb{E}[\delta_{kg}|e_g]) = \left(\frac{\phi_{kg}}{\pi_k}\right)^2 \operatorname{Var}(e_g) = \phi_{kg} \tau_{kg}^2, \quad \Rightarrow \quad R^2 = \frac{\phi_{kg} \tau_{kg}^2}{\tau_{kg}^2} = \phi_{kg}.
\end{equation}

Thus $R^2 = \text{corr}^2(\hat{\delta}_{kg},\delta_{kg})=\phi_{kg}$ under Gaussian MMSE. For any predictor $f(e_g)$, $\operatorname{Cov}(f(e_g),\delta_{kg})=\operatorname{Cov}(f(e_g),\mathbb{E}[\delta_{kg}\mid e_g])$; Cauchy--Schwarz therefore bounds its squared correlation by $\phi_{kg}$. Since $R^2\le\mathrm{corr}^2$, the stated inequality follows.

\textbf{Mutual information.} At fixed composition, $(\delta_{kg},e_g)$ is jointly Gaussian with squared correlation $\phi_{kg}$ (Step~3). A Gaussian of variance $s^2$ has entropy $\tfrac12\log(2\pi e\,s^2)$, and $\mathrm{Var}(\delta_{kg}\mid e_g,\bm{\pi})=\tau_{kg}^2(1-\phi_{kg})$, so
\begin{equation}
  I(\delta_{kg};e_g\mid\bm{\pi}=\bm{p}) = -\tfrac12\log\bigl(1-\phi_{kg}(\bm{p})\bigr),
\end{equation}
the standard identity for jointly Gaussian scalars~\cite{cover2006elements}; the same holds with $\phi^{\mathrm{CT}}_{kg}$ for general $\bm{\Sigma}_g$.

\textbf{General case} (arbitrary $\bm{\Sigma}_g$): The Gaussian MMSE is $\mathbb{E}[\bm{\delta}_{\cdot g}|e_g] = \frac{\bm{\Sigma}_g\bm{\pi}}{\bm{\pi}^{\!\top}\bm{\Sigma}_g\bm{\pi}} e_g$, yielding
\begin{equation}
R^2(\delta_{kg}) = \frac{[(\bm{\Sigma}_g\bm{\pi})_k]^2}{[\bm{\Sigma}_g]_{kk} \cdot (\bm{\pi}^{\!\top}\bm{\Sigma}_g\bm{\pi})} = \phi^{\mathrm{CT}}_{kg}.
\end{equation}
Cross-cell-type correlations enter via $\bm{\Sigma}_g\bm{\pi}$: when cell types have correlated expression, the second term in the numerator sum $(\bm{\Sigma}_g\bm{\pi})_k = [\bm{\Sigma}_g]_{kk}\pi_k + \sum_{k'\neq k}[\bm{\Sigma}_g]_{kk'}\pi_{k'}$ can amplify or reduce signal. Since $\mu_{kg} = \mu_{kg}^* + \delta_{kg}$, the result holds for $\mu_{kg}$ (correlation is translation-invariant).

\textbf{Remark (simultaneity).} The MMSE estimators achieve the ceiling for all $K$ cell types simultaneously. They are the Gaussian conditional expectations $\mathbb{E}[\delta_{kg} | e_g]$, each optimal, and jointly satisfy the consistency constraint $\sum_k \pi_k \hat{\delta}_{kg} = e_g$. Thus there is no trade-off: achieving the ceiling for one cell type does not prevent achieving it for others.

\textbf{Varying compositions.} The steps above hold at a fixed composition $\bm{\pi}$, where the conditional-mean estimator has error variance $\tau_{kg}^2\bigl(1-\phi_{kg}(\bm{\pi})\bigr)$. If the deviations are independent of composition, so that $\mathrm{Var}(\delta_{kg})=\tau_{kg}^2$ for every $\bm{\pi}$, averaging this error over samples gives a pooled mean squared error $\tau_{kg}^2\bigl(1-\mathbb{E}_i[\phi_{kg}(\bm{\pi}^{(i)})]\bigr)$, hence a pooled $R^2 = \mathbb{E}_i[\phi_{kg}(\bm{\pi}^{(i)})]$; since the estimator is a conditional mean, its squared correlation with $\delta_{kg}$ equals this $R^2$. The same holds with $\phi^{\mathrm{CT}}_{kg}$ for general $\bm{\Sigma}_g$. Because $\phi_{kg}$ is nonlinear in $\bm{\pi}$, the plug-in $\phi_{kg}(\bar{\bm{\pi}})$ used in our figures differs from this average by a Jensen gap of either sign.

\textbf{Population-level interpretation.} Equation~\eqref{eq:r2} gives conditional performance at fixed composition; with varying compositions, the conditional mean $\mathbb{E}[\delta_{kg}\mid e_g,\bm{\pi}]$ attains pooled $R^2=\mathbb{E}_i[\phi_{kg}(\bm{\pi}^{(i)})]$ under the stated assumptions. Our figures instead use mean-composition plug-in scores. Their correlations with the MLP's log-space test $R^2$ ($0.49$ diagonal; $0.39$ cross-CT) support their use as empirical difficulty measures, rather than establishing absolute ceilings for that model.

\textbf{Measurement noise and co-variation.} With additional independent, zero-mean assay noise of variance $\sigma_g^2$, the best linear score becomes $[(\bm{\Sigma}_g\bm{\pi})_k]^2/\bigl([\bm{\Sigma}_g]_{kk}(\bm{\pi}^{\!\top}\bm{\Sigma}_g\bm{\pi}+\sigma_g^2)\bigr)$, the form underlying Unico~\cite{chen2025unico}. It remains a ceiling for nonlinear per-gene estimators if the deviations and assay noise are jointly Gaussian. Additional noise lowers this score for a fixed biological distribution. If expression co-varies with composition, $\bm{\pi}$ alone can predict $\mu_{kg}$; the calculation must then use $\mathbb{E}[\delta_{kg}\mid\bm{\pi}]$ and $\mathrm{Cov}(\bm{\delta}_g\mid\bm{\pi})$ (cf.\ App.~\ref{app:varrho:covary}).

\subsection{Estimators Benchmarked in DALE-Eval}
\label{app:dale_methods}

DALE-Eval~\cite{hu2025daleeval} benchmarks seven CTSE methods, with proportions estimated by InstaPrism (a fast reimplementation of BayesPrism) supplied to those that require them. Table~\ref{tab:dale_methods} classifies their estimators given proportions and fitted parameters. Five are per-gene linear estimators; Proposition~\ref{prop:ceiling_intro} characterizes their oracle linear-in-CPM versions under the assumed model. TCA and Unico use the Gaussian conditional-mean form; bMIND averages such estimates over parameter draws. With oracle proportions and correct fixed parameters, the conditional-mean estimator attains the relevant ceiling. This statement does not automatically extend to the complete fitted algorithms: bulk-estimated proportions and cohort-fitted parameters can carry information from other genes or samples.

\begin{table}[h]
\centering
\small
\caption{\textbf{CTSE estimators benchmarked in DALE-Eval}, given $\hat{\bm{\pi}}$ and fitted parameters, in our notation. $a_{kg}$: reference probability that a read from CT $k$ comes from gene $g$ ($\sum_g a_{kg}=1$); $\mu^*_{kg}$, $\tau^2_{kg}$, $\bm{\Sigma}_g$: CT means, across-sample variances and $K{\times}K$ covariance at gene $g$; $\sigma_g^2$: measurement-noise variance. TCA and Unico fit these on the bulk cohort; bMIND uses single-cell priors updated on bulk.}
\label{tab:dale_methods}
\begin{tabular}{@{}p{2.4cm}p{6.9cm}cc@{}}
\toprule
Method & Estimate of $\mu_{kg}^{(i)}$ & per-gene & linear in $x_g$ \\
\midrule
InstaPrism / BayesPrism~\cite{hu2024instaprism,chu2022bayesprism} & mean read allocation $Z_{gk}^{(i)} = x_g^{(i)}\hat\pi_k a_{kg}\big/\sum_{k'}\hat\pi_{k'}a_{k'g}$, normalised within CT $k$: $\hat\mu_{kg}^{(i)} = x_g^{(i)}a_{kg}\big/\sum_{k'}\hat\pi_{k'}a_{k'g}$ ($\hat{\bm{\pi}}$ = read fractions, i.e.\ $\bm{\pi}^{\mathrm{lib}}$). Conditional mean of the multinomial read allocation in BayesPrism's model (its Eqs.~3--6, 10--11); InstaPrism computes this mean in place of Gibbs sampling (its Sec.~2.3) & \checkmark & \checkmark \\
TCA~\cite{rahmani2019} & $\mu^*_{kg} + \hat\pi_k\tau^2_{kg}\bigl(x_g - \hat{\bm{\pi}}^{\!\top}\bm{\mu}^*_g\bigr)\big/\bigl(\sum_{k'}\hat\pi_{k'}^2\tau^2_{k'g}+\sigma_g^2\bigr)$, i.e.\ Unico with diagonal $\bm{\Sigma}_g$ (its Eqs.~5--8, by Woodbury); $R^2 = \phi_{kg}$ at $\sigma_g^2=0$ & \checkmark & \checkmark \\
Unico~\cite{chen2025unico} & $\bm{\mu}^*_g + \bm{\Sigma}_g\hat{\bm{\pi}}\bigl(x_g - \hat{\bm{\pi}}^{\!\top}\bm{\mu}^*_g\bigr)\big/\bigl(\hat{\bm{\pi}}^{\!\top}\bm{\Sigma}_g\hat{\bm{\pi}}+\sigma_g^2\bigr)$ (Theorem~1); $R^2 = \phi^{\mathrm{CT}}_{kg}$ at $\sigma_g^2=0$ & \checkmark & \checkmark \\
bMIND~\cite{wang2021bmind} & model $x_g = \hat{\bm{\pi}}^{\!\top}\bm{\mu}_g + e_g$, $\bm{\mu}_g\sim\mathcal{N}(\bm{\mu}^*_g,\bm{\Sigma}_g)$ (its Eq.~1), with single-cell priors on $\bm{\mu}^*_g$, $\bm{\Sigma}_g$; posterior mean = Unico's form, averaged over MCMC draws of $(\bm{\mu}^*_g,\bm{\Sigma}_g,\sigma_g^2)$ & \checkmark & \checkmark$^{\dagger}$ \\
EPIC-unmix~\cite{tang2025epicunmix} & bMIND's model (its Eq.~1) fitted twice: the second MCMC run re-centres the priors on the first run's per-gene estimates for the target samples; posterior mean = Unico's form, averaged over the second run's draws & \checkmark & \checkmark$^{\dagger}$ \\
CIBERSORTx (high-res.)~\cite{newman2019cibersortx} & per gene: sort samples by $x_g$, fit non-negative least squares on sample groups, test and smooth over a sliding window & \checkmark & \xmark \\
ENIGMA~\cite{wang2023enigma} & $\arg\min_{\{\bm{X}_k\}} \tfrac{\alpha}{2}\|\bm{O}-\sum_k\bm{X}_k\mathrm{diag}(\bm{\theta}_k)\|_F^2 + \tfrac{1-\alpha}{2}\sum_k\|\bm{X}_k\bm{a}-\bm{r}_k\|^2 + \beta\sum_k \Omega(\bm{X}_k)$ (its Eqs.~1--3), with $\bm{X}_k$ genes${\times}$samples and $\Omega$ the trace norm or maximum row $\ell_2$ norm & \xmark & \xmark \\
\bottomrule
\end{tabular}

\smallskip
\raggedright\footnotesize $^{\dagger}$Applied to log-transformed bulk in DALE-Eval, so linear in $\log x_g$.
\end{table}

\textbf{Caveats.} (i) bMIND and EPIC-unmix were run on log-transformed bulk, so the linear-CPM ceiling does not directly describe those runs (App.~\ref{app:notation:log}). (ii) Substituting estimated proportions into $\phi_{kg}$ does not generally produce a valid new ceiling: composition errors alter the predictor, and estimated proportions can encode cross-gene information. (iii) A nonlinear per-gene estimator falls under the ceiling only with the Gaussian assumptions and the stated oracle inputs. ENIGMA couples genes through its regulariser and lies outside this class, like our MLP.

\subsection{Why Some Targets Are Harder}
\label{app:harder}

Under independent CT variation, $\phi_{kg}\in[0,1]$ measures the fraction of residual bulk variance contributed by CT $k$ at gene $g$. A small share can arise from low abundance or small within-CT variance relative to the other CTs. Figures~\ref{fig:phi_diag} and~\ref{fig:phi_ct} show the full CT-by-target-gene grid, with the 106 selected pairs marked. Colour uses the square root of the score to spread small values; the linear $R^2$ ceiling itself is the untransformed score.

\subsection{Cross-CT Covariance and Estimation}
\label{app:cross_ct}

\begin{figure}[tb]
\centering
\includegraphics[width=0.6\textwidth]{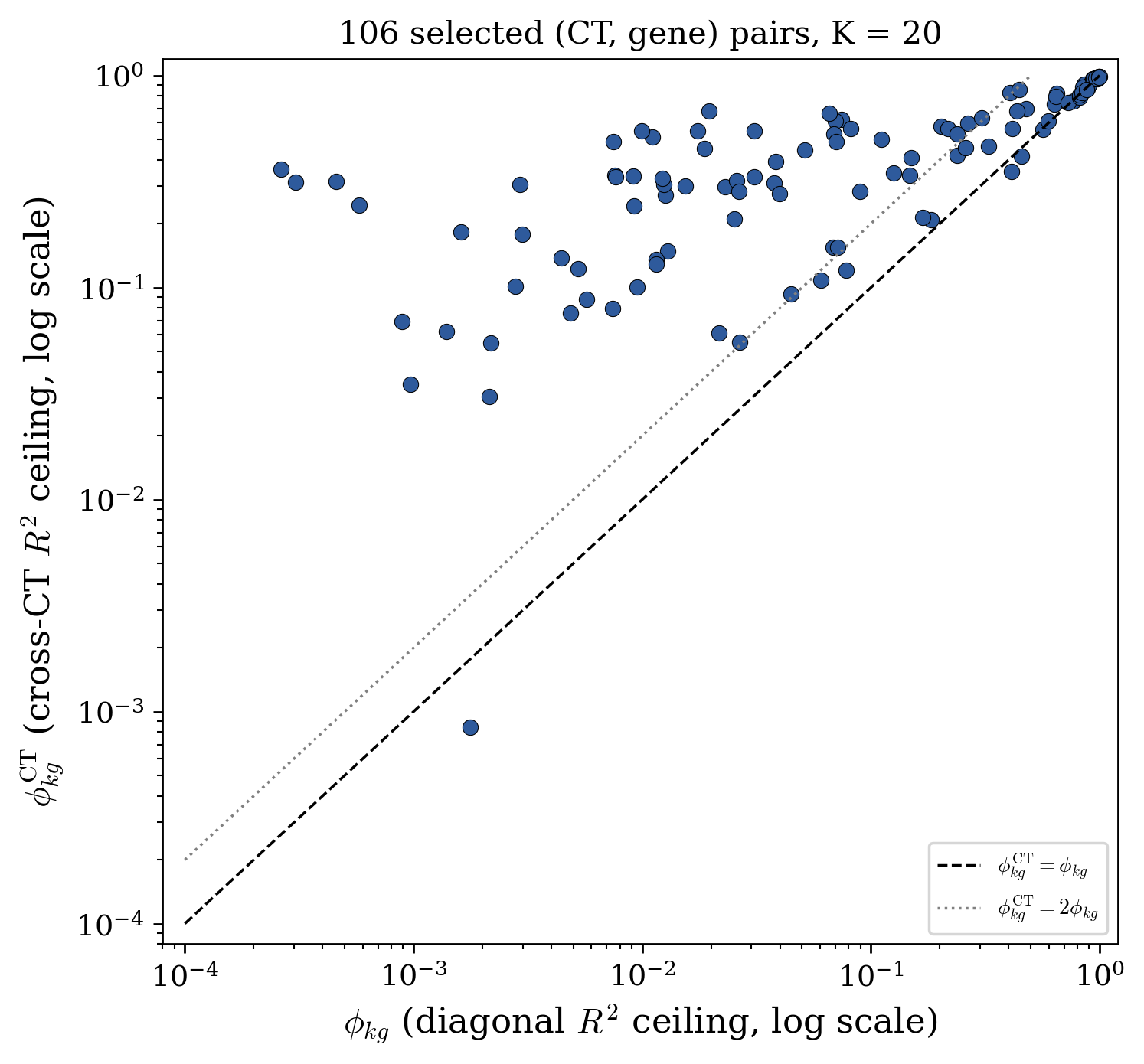}
\caption{\textbf{Cross-CT covariance changes per-gene learnability.} $\phi_{kg}$ (independent CTs) vs $\phi^{\mathrm{CT}}_{kg}$ (same-gene cross-CT covariance) for the 106 selected pairs, on log scales. Points above the dashed diagonal have a higher ceiling under the correlated model; the dotted line marks a twofold increase. Low-signal pairs show the largest relative gains. Cross-CT covariance changes the ceiling, but the predictor remains per-gene.}
\label{fig:phi_ct_vs_diag}
\end{figure}

\textbf{Generalized ceiling for per-gene estimators.} The diagonal bound $\phi_{kg}$ assumes independence across cell types at gene $g$ (diagonal $\bm{\Sigma}_g$). When cell types have correlated expression at gene $g$, the ceiling is $\phi^{\mathrm{CT}}_{kg}$ (Eq.~\eqref{eq:phi_ct}), the $R^2$ of the best linear per-gene predictor under the true $K{\times}K$ covariance. Writing $\tau_{kg}^2 = [\bm{\Sigma}_g]_{kk}$ and $c_{kk'g} = [\bm{\Sigma}_g]_{kk'}$,
\begin{equation}
\phi^{\mathrm{CT}}_{kg} = \frac{\bigl(\pi_k\tau_{kg}^2 + \sum_{k'\neq k}\pi_{k'}c_{kk'g}\bigr)^2}{\tau_{kg}^2\bigl(\sum_j \pi_j^2\tau_{jg}^2 + \sum_{j\neq l}\pi_j\pi_l c_{jlg}\bigr)},
\end{equation}
which reduces to $\phi_{kg}$ when all $c_{kk'g}=0$. Cross-covariances enter both numerator and denominator, so $\phi^{\mathrm{CT}}_{kg}$ need not exceed $\phi_{kg}$. Positive covariance with the target CT can provide signal, whereas covariance among other CTs can add interference. Negative covariance can cancel target signal: at $K{=}2$, equal proportions and variances give $\phi^{\mathrm{CT}}=(1+\rho)/2$, below $\phi=1/2$ when $\rho<0$. Most atlas combinations show higher correlated scores (Fig.~\ref{fig:phi_ct_vs_diag}), consistent with shared sample or microenvironment effects. The observation remains one gene plus composition; the estimator exploits a cross-CT prior, not additional measured genes.

\textbf{Covariance structure.} Stacking CT deviations gives a $KG{\times}KG$ covariance $\bm{C}$. Per-gene ceilings use its $K{\times}K$ same-gene blocks; pooling $M$ genes would use $MK{\times}MK$ sub-blocks. At fixed composition, the bulk residual covariance is $\bm{H}\bm{C}\bm{H}^{\!\top}$, where $\bm{H}=[\pi_1\bm{I}_G,\ldots,\pi_K\bm{I}_G]$. Under composition-independent deviations, the proportion result uses its average over compositions, $\bm{T}_e=\mathbb{E}_{\bm{\pi}}[\bm{H}\bm{C}\bm{H}^{\!\top}]$ (App.~\ref{app:varrho:main}).

\textbf{Estimation feasibility.} A per-gene $K{\times}K$ covariance has 210 parameters at $K{=}20$, making it more tractable than full gene covariance. The atlas contains 556 samples from 318 donors; precision depends on CT overlap, repeated donors and heavy tails, not just the sample-to-parameter ratio. Pairwise-complete covariance estimates need not be positive semidefinite; we therefore clip negative eigenvalues to zero and add a $10^{-6}$ diagonal ridge before computing scores, which use the mean library-weighted composition of the pseudobulk training split. At $G\approx1000$, unregularized gene-covariance inversion is underdetermined at this sample size; structured or regularized estimation is a possible extension. We therefore report diagonal and cross-CT plug-in scores, rather than a whole-profile ceiling.

\subsection{Learnability and Prognostic Signal}
Across the 106 selected pairs, the plug-in $\phi^{\mathrm{CT}}_{kg}$ shows no clear rank association with $|\log\mathrm{HR}|$ per SD or C-index gain: Spearman $\rho=0.02$ ($p=0.82$) and $0.03$ ($p=0.73$), respectively (Fig.~\ref{fig:combined}). This is an exploratory comparison within a curated, correlated target set, not evidence of statistical independence or a general absence of a relationship.

\subsection{Geometric Interpretation and Variance-Share Distribution}
\label{app:geom_interp}

\begin{figure}[tb]
\centering
\resizebox{0.92\textwidth}{!}{\begin{tikzpicture}[x=0.98cm,y=0.64cm,>={Stealth[length=3.5mm,width=2.6mm]},
  lab/.style={font=\fontsize{22}{26}\selectfont},
  sm/.style={font=\fontsize{18}{21}\selectfont},
  bg/.style={fill=white,fill opacity=0.85,text opacity=1,inner sep=1.5pt,rounded corners=1mm}]
  \pgfmathsetseed{11}

  \draw[->,line width=1.6pt,muted] (0,0) -- (25,0) node[below left,sm,text=muted,yshift=-2mm] {gene $g$};
  \draw[->,line width=1.6pt,muted] (0,0) -- (0,15.5) node[above right,sm,text=muted] {gene $g'$};

  \coordinate (m1) at (4.6,3.4);                 %
  \coordinate (m2) at (18.6,9.0);              %
  \coordinate (s1) at ($(m1)+(1.4,-0.2)$);     %
  \coordinate (s2) at ($(m2)+(-0.36,2.1)$);    %
  \coordinate (mbar) at ($(m2)!0.4!(m1)$);     %
  \coordinate (x) at ($(s2)!0.4!(s1)$);        %

  \foreach \i in {1,...,110} {
    \pgfmathsetmacro{\u}{1.4*(rand+rand+rand)}
    \pgfmathsetmacro{\v}{0.33*(rand+rand+rand)}
    \pgfmathsetmacro{\a}{\u*cos(14.64)-\v*sin(14.64)}\pgfmathsetmacro{\b}{\u*sin(14.64)+\v*cos(14.64)}
    \fill[ctA!70] ($(m1)+(\a cm,\b cm)$) circle (2.2pt);
  }
  \foreach \i in {1,...,160} {
    \pgfmathsetmacro{\u}{1.2*(rand+rand+rand)}
    \pgfmathsetmacro{\v}{0.5*(rand+rand+rand)}
    \pgfmathsetmacro{\a}{\u*cos(104.64)-\v*sin(104.64)}\pgfmathsetmacro{\b}{\u*sin(104.64)+\v*cos(104.64)}
    \fill[ctB!70] ($(m2)+(\a cm,\b cm)$) circle (2.2pt);
  }

  \draw[ctA!60!black,densely dotted,line width=1.2pt] (m1) -- ($(m1)+(0,2.7)$);
  \draw[->,ctA!60!black,line width=2.2pt] ($(m1)+(0,2.5)$) -- ($(m1)+(1.385,2.5)$)
    node[midway,above,lab,text=ctA!60!black] {$\tau_{1g}$};
  \draw[ctB!55!black,densely dotted,line width=1.2pt] (m2) -- ($(m2)+(0,6.2)$);
  \draw[->,ctB!55!black,line width=2.2pt] ($(m2)+(0,6.0)$) -- ($(m2)+(0.583,6.0)$)
    node[midway,above,lab,text=ctB!55!black] {$\tau_{2g}$};

  \draw[dashed,line width=1.4pt,black!45,rotate around={-12:(mbar)}] (mbar) ellipse (2.0 and 1.15);

  \draw[->,dashed,line width=2pt,navy!80] (m2) -- (m1);
  \draw[line width=1.8pt,black!45] (s1) -- (s2);

  \draw[->,ctA!60!black,line width=2.2pt,densely dashed] (m1) -- (s1);
  \draw[->,ctB!55!black,line width=2.2pt,densely dashed] (m2) -- (s2);

  \node[ctA!60!black,font=\fontsize{26}{26}\selectfont] at (m1) {$\times$};
  \node[ctB!55!black,font=\fontsize{26}{26}\selectfont] at (m2) {$\times$};
  \fill[ctA!60!black] (s1) circle (3.2mm);
  \fill[ctB!55!black] (s2) circle (3.2mm);
  \fill[black!55] (mbar) circle (2.6mm);
  \draw[->,line width=2.6pt,black] (mbar) -- (x);
  \fill[black] (x) circle (3.6mm);

  \node[lab,text=ctA!60!black,anchor=east,bg] at ($(m1)+(-0.9,0.45)$) {$\bm{\mu}^*_1$};
  \node[lab,text=ctA!60!black,anchor=north west,bg] at ($(s1)+(0.45,-0.35)$) {$\bm{\mu}^{(i)}_1$};
  \node[lab,text=ctA!60!black,anchor=north east,bg] at ($(m1)+(0.5,-0.6)$) {$\bm{\delta}^{(i)}_1$};
  \node[lab,text=ctB!55!black,anchor=north west,bg] at ($(m2)+(0.4,-0.3)$) {$\bm{\mu}^*_2$};
  \node[lab,text=ctB!55!black,anchor=east,bg] at ($(s2)+(-0.5,0.4)$) {$\bm{\mu}^{(i)}_2$};
  \node[lab,text=ctB!55!black,anchor=west,bg] at ($(m2)!0.5!(s2)+(0.4,0.1)$) {$\bm{\delta}^{(i)}_2$};
  \node[lab,text=navy,anchor=north,bg] at ($(m2)!0.68!(m1)+(0.3,-0.45)$) {$\bm{\Delta}$};
  \node[lab,anchor=east,bg] at ($(x)+(-0.7,0.6)$) {$\bm{x}^{(i)}$};
  \node[lab,anchor=west,bg] at ($(mbar)!0.5!(x)+(0.55,0.45)$) {$\bm{e}^{(i)}$};
  \node[sm,text=black!80,anchor=north west,bg] at ($(mbar)+(0.4,-1.0)$) {$\sum_k\pi_k\bm{\mu}^*_k$};
  \node[sm,text=black!80,anchor=north east,bg] at ($(mbar)+(-1.6,-1.3)$) {$\bm{T}_e$};

  \begin{scope}[shift={(2.6,10.6)}]
    \coordinate (o) at (0,0);
    \coordinate (a) at ($2*0.4*(1.4,-0.2)$);
    \coordinate (b) at ($(a)+2*0.6*(-0.36,2.1)$);
    \draw[->,ctA!60!black,line width=2.4pt] (o) -- (a);
    \draw[->,ctB!55!black,line width=2.4pt] (a) -- (b);
    \draw[->,black,line width=2.6pt] (o) -- (b);
    \node[sm,text=ctA!60!black,anchor=north] at ($(o)!0.5!(a)+(0,-0.15)$) {$\pi_1\bm{\delta}^{(i)}_1$};
    \node[sm,text=ctB!55!black,anchor=west] at ($(a)!0.5!(b)+(0.25,0)$) {$\pi_2\bm{\delta}^{(i)}_2$};
    \node[sm,anchor=east] at ($(o)!0.55!(b)+(-0.2,0)$) {$\bm{e}^{(i)}$};
  \end{scope}
\end{tikzpicture}%
}
\caption{\textbf{Geometry of the bulk mixture residual} ($K{=}2$, two genes; schematic). Dots are sample-specific profiles $\bm{\mu}_k^{(i)}$ around the centroids $\bm{\mu}_k^*$ ($\times$); $\tau_{kg}$ is their spread along gene $g$. The bulk point $\bm{x}^{(i)}=\sum_k\pi_k\bm{\mu}_k^{(i)}$ lies between the sample-specific CT profiles, while the centroid mixture $\sum_k\pi_k\bm{\mu}_k^*$ (grey dot) uses their population means at the same composition. Their difference is $\bm{e}=\sum_k\pi_k\bm{\delta}_k$ (inset), with covariance $\bm{T}_e$ (dashed ellipse). Variation of CT 1 along the contrast $\bm{\Delta}$ makes $\pi$ harder to read; variation of CT 2 perpendicular to it does not (App.~\ref{app:varrho:main}). At fixed composition and independent CT variation, the residual variance at a gene partitions into CT contributions; $\phi_{kg}$ is CT $k$'s share. The diagram illustrates mixture geometry, not a whole-profile accuracy ceiling.}
\label{fig:tau_phi}
\end{figure}
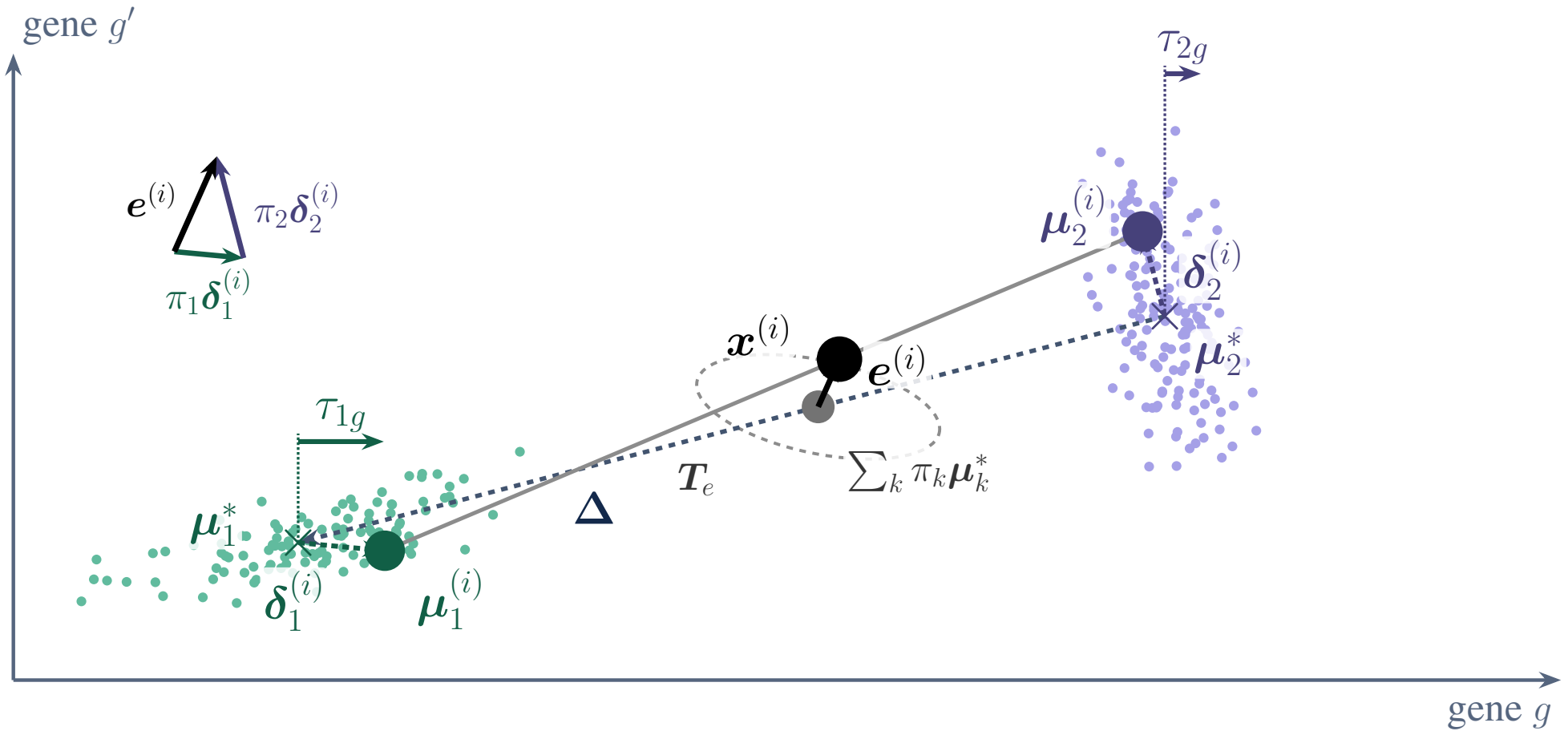

Figure~\ref{fig:tau_phi} illustrates the mixture residual underlying $\phi_{kg}$; Figure~\ref{fig:phi_ct_vs_diag} shows how cross-CT covariance changes learnability.

\begin{figure}[tb]
\centering
\includegraphics[width=0.9\textwidth]{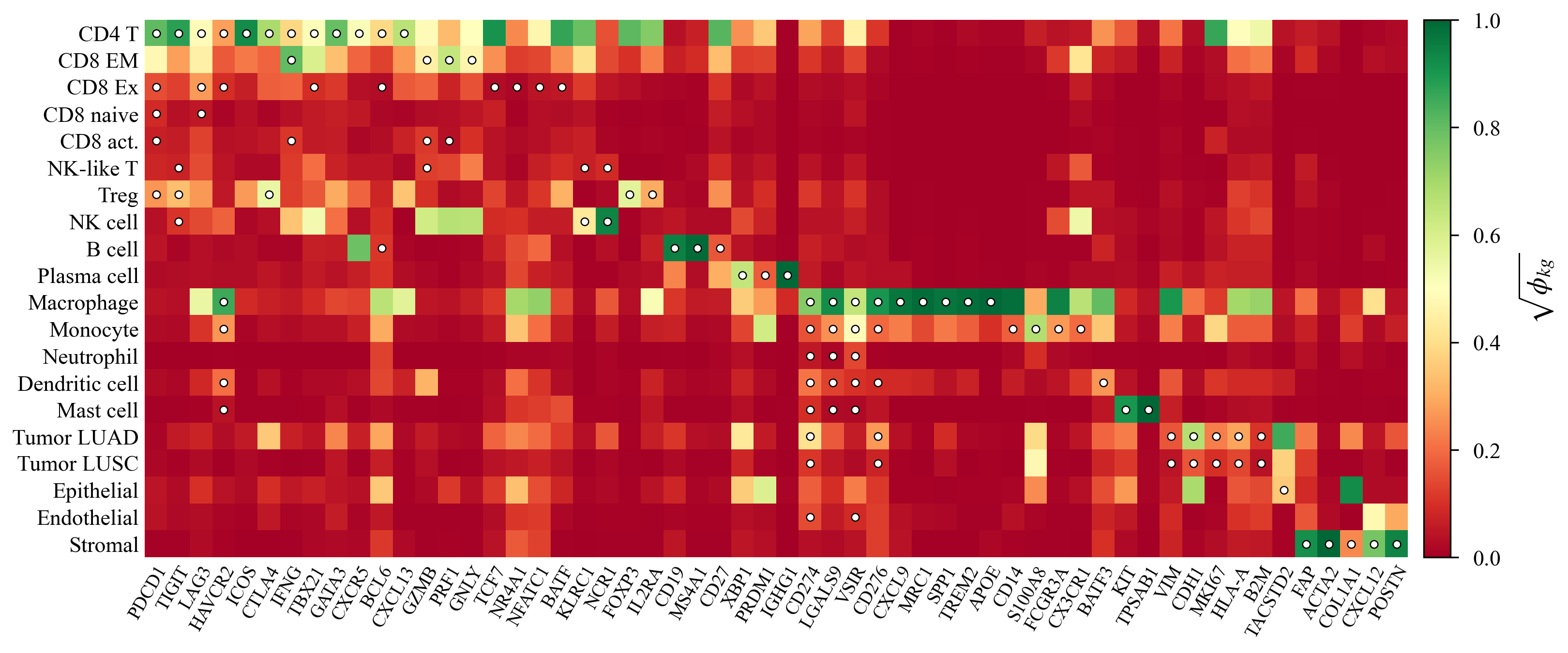}
\caption{\textbf{Independent-CT variance shares across 20 CTs and 56 target genes.} Color shows $\sqrt{\phi_{kg}}$ at mean composition under diagonal $\bm{\Sigma}_g$. Dots mark the 106 selected pairs. Abbreviations: EM, effector memory; Ex, exhausted; act., activated; Treg, regulatory T cells. Higher values indicate greater target contribution to residual bulk variance; the untransformed $\phi_{kg}$ is the linear $R^2$ ceiling under these assumptions.}
\label{fig:phi_diag}
\end{figure}

\textbf{Signal distribution across targets.} Under the independent-CT plug-in, 49 of 106 selected pairs have $\phi_{kg}<0.05$ (Fig.~\ref{fig:phi_diag}). Accounting for same-gene cross-CT covariance changes these scores (Fig.~\ref{fig:phi_ct}, main text). Whole-profile models can use additional cross-gene information and are outside either per-gene ceiling.

\clearpage
\section{Proportion Estimation: Proofs and Details}
\label{app:varrho}

At fixed composition, $\phi^{\mathrm{CT}}_{kg}$ is the population $R^2$ of the best linear per-gene predictor of $\mu_{kg}$; it reduces to $\phi_{kg}$ under independent CT variation. Proposition~\ref{prop:varrho} gives the analogous optimal linear accuracy for proportions with two CTs; this appendix gives the general-$K$ result and proof.

\subsection{Setup, general-$K$ result and proof}
\label{app:varrho:main}

Let $\bm{\delta}_k := \bm{\mu}_k - \bm{\mu}_k^* \in \mathbb{R}^G$ be the per-sample deviations, with within-CT gene covariance $\bm{T}_k := \mathrm{Cov}(\bm{\delta}_k) \in \mathbb{R}^{G\times G}$. Since $\sum_k \pi_k = 1$, only $K{-}1$ proportions are free. Taking CT $K$ as reference, write $\tilde{\bm{\pi}} := (\pi_1,\ldots,\pi_{K-1})^{\!\top} \in \mathbb{R}^{K-1}$, so that $\pi_K = 1 - \bm{1}^{\!\top}\tilde{\bm{\pi}}$, and Eq.~\eqref{eq:exact_lin} becomes
\begin{equation}
\label{eq:x_contrast}
\begin{gathered}
  \bm{x} = \bm{\mu}_K^* + \bm{D}\,\tilde{\bm{\pi}} + \bm{e}, \\
  \bm{D} := \bigl[\bm{\mu}_1^*-\bm{\mu}_K^*,\ \ldots,\ \bm{\mu}_{K-1}^*-\bm{\mu}_K^*\bigr] \in \mathbb{R}^{G\times(K-1)}, \qquad
  \bm{e} := \sum_{k=1}^K \pi_k\,\bm{\delta}_k \in \mathbb{R}^G ,
\end{gathered}
\end{equation}
where $\bm{e}$ is the vector of bulk residuals $e_g$ defined in Prop.~\ref{prop:ceiling_intro}. Let $\bm{V} := \mathrm{Cov}(\tilde{\bm{\pi}}) \in \mathbb{R}^{(K-1)\times(K-1)}$ be the covariance of the free coordinates (the $K{\times}K$ matrix $\mathrm{Cov}(\bm{\pi})$ is singular), and $\bm{T}_e := \mathrm{Cov}(\bm{e}) \in \mathbb{R}^{G\times G}$. When the deviations are independent of composition, $\bm{T}_e = \sum_{k,k'}\mathbb{E}[\pi_k\pi_{k'}]\,\mathrm{Cov}(\bm{\delta}_k,\bm{\delta}_{k'})$: a population quantity that depends on the composition distribution only through its second moments, not on any particular sample's $\bm{\pi}$. Given $\bm{\pi}$, the residual covariance does vary with $\bm{\pi}$; only nonlinear predictors can exploit this (App.~\ref{app:varrho:nongauss}).

\textbf{Proposition~\ref{prop:varrho} (proportion estimation), general $K$.} Assume $\mathrm{Cov}(\bm{e},\tilde{\bm{\pi}}) = \bm{0}$, which holds if the deviations are mean-independent of composition, $\mathbb{E}[\bm{\delta}_k \mid \bm{\pi}] = \bm{0}$ for all $k$, and $\bm{V}, \bm{T}_e \succ 0$. Then the best linear predictor of $\tilde{\bm{\pi}}$ from $\bm{x}$ has error covariance
\begin{equation}
\label{eq:pi_err_cov}
  \bm{E} = \bigl(\bm{V}^{-1} + \bm{P}\bigr)^{-1} \in \mathbb{R}^{(K-1)\times(K-1)}, \qquad
  \bm{P} := \bm{D}^{\!\top}\bm{T}_e^{-1}\bm{D},
\end{equation}
the per-CT proportion $R^2$ is
\begin{equation}
\label{eq:eta_k}
  \eta_k := 1 - \frac{E_{kk}}{V_{kk}} \quad (k<K), \qquad
  \eta_K := 1 - \frac{\bm{1}^{\!\top}\bm{E}\,\bm{1}}{\mathrm{Var}(\pi_K)},
\end{equation}
and these values do not depend on the choice of reference CT. For $K{=}2$ this reduces to Proposition~\ref{prop:varrho}, and $\varrho \ge \varrho_\parallel := \|\bm{\Delta}\|/\sqrt{s_1^2+s_2^2}$, where $s_k^2 := \bm{u}^{\!\top}\bm{T}_k\bm{u}$ is the variance of CT $k$ along the contrast direction $\bm{u} := \bm{\Delta}/\|\bm{\Delta}\|$. Hence the mean squared error of the best linear predictor of $\pi$, and therefore of the Bayes-optimal predictor, is at most $1/\varrho_\parallel^2$.

\begin{figure}[h]
\centering
\scalebox{0.75}{\begin{tikzpicture}
\begin{scope}
\node[anchor=west,font=\bfseries\scriptsize] at (-0.1,3.69) {(a) tight clouds};
\draw[gray!55,line width=0.4pt] (0,0) -- (3.91,0);
\draw[gray!55,line width=0.4pt] (0,0) -- (0,3.53);
\begin{scope}
\clip (-0.02,-0.02) rectangle (3.94,3.56);
\fill[cone,opacity=0.13] (0.806,2.728) -- ++(-43.5:7.2) arc[start angle=-43.5,end angle=-27.6,radius=7.2] -- cycle;
\draw[cB,opacity=0.55,dashed,line width=0.4pt,rotate around={-36.87:(0.806,2.728)}] (0.806,2.728) ellipse (0.310 and 0.310);
\fill[cB,opacity=0.11,rotate around={-36.87:(0.806,2.728)}] (0.806,2.728) ellipse (0.310 and 0.310);
\draw[cA,opacity=0.55,dashed,line width=0.4pt,rotate around={-36.87:(3.038,1.054)}] (3.038,1.054) ellipse (0.310 and 0.310);
\fill[cA,opacity=0.11,rotate around={-36.87:(3.038,1.054)}] (3.038,1.054) ellipse (0.310 and 0.310);
\draw[seg,line width=0.45pt] (0.583,2.638) -- (3.068,1.342);
\draw[seg,line width=0.45pt] (0.814,2.803) -- (2.902,0.820);
\draw[seg,line width=0.45pt] (0.957,2.636) -- (2.919,0.918);
\draw[seg,line width=0.45pt] (0.664,2.682) -- (2.915,1.052);
\draw[seg,line width=0.45pt] (1.051,2.861) -- (3.001,1.189);
\draw[seg,line width=0.45pt] (0.725,2.550) -- (2.975,0.987);
\draw[cstar,line width=1.1pt] (0.806,2.728) -- (3.038,1.054);
\fill[cB] (0.806,2.728) circle (2.1pt); \fill[cA] (3.038,1.054) circle (2.1pt);
\draw[xpt,line width=1.1pt,{Stealth[length=3.5pt]}-{Stealth[length=3.5pt]}] (2.683,0.839) -- (2.931,0.653);
\draw[xpt!45,line width=0.35pt,dotted] (2.914,1.147) -- (2.683,0.839);
\draw[xpt!45,line width=0.35pt,dotted] (3.162,0.961) -- (2.931,0.653);
\end{scope}
\node[anchor=east,font=\scriptsize] at (0.632,2.914) {$\bm{\mu}_2^*$};
\node[anchor=west,font=\scriptsize] at (3.410,1.271) {$\bm{\mu}_1^*$};
\node[anchor=north,font=\tiny,xpt] at (2.688,0.588) {$s_1$};
\node[font=\scriptsize,anchor=north] at (1.92,-0.34) {narrow cone};
\node[font=\scriptsize,anchor=north] at (1.92,-0.96) {$\pi$ well determined};
\draw[cstar,line width=0.9pt,-{Stealth[length=3.5pt]}] (0.806,2.728) -- (1.376,2.300);
\node[anchor=south west,font=\tiny] at (1.389,2.313) {$\bm{u}$};
\draw[cstar,line width=0.7pt,{Stealth[length=3pt]}-{Stealth[length=3pt]}] (0.962,2.936) -- (3.194,1.262);
\node[font=\tiny,anchor=south] at (2.138,2.179) {$\Delta$};
\draw[cA,line width=0.7pt,{Stealth[length=3pt]}-] (3.038,1.054) -- (3.193,1.054);
\node[font=\tiny,cA,anchor=west] at (3.243,1.116) {$\tau$};
\end{scope}
\begin{scope}[xshift=5.2cm]
\node[anchor=west,font=\bfseries\scriptsize] at (-0.1,3.69) {(b) spread $\perp$ to $\bm{u}$};
\draw[gray!55,line width=0.4pt] (0,0) -- (3.91,0);
\draw[gray!55,line width=0.4pt] (0,0) -- (0,3.53);
\begin{scope}
\clip (-0.02,-0.02) rectangle (3.94,3.56);
\fill[cone,opacity=0.13] (0.806,2.728) -- ++(-53.2:7.2) arc[start angle=-53.2,end angle=-31.9,radius=7.2] -- cycle;
\draw[cB,opacity=0.55,dashed,line width=0.4pt,rotate around={53.13:(0.806,2.728)}] (0.806,2.728) ellipse (1.178 and 0.273);
\fill[cB,opacity=0.11,rotate around={53.13:(0.806,2.728)}] (0.806,2.728) ellipse (1.178 and 0.273);
\draw[cA,opacity=0.55,dashed,line width=0.4pt,rotate around={53.13:(3.038,1.054)}] (3.038,1.054) ellipse (1.178 and 0.273);
\fill[cA,opacity=0.11,rotate around={53.13:(3.038,1.054)}] (3.038,1.054) ellipse (1.178 and 0.273);
\draw[seg,line width=0.45pt] (0.667,2.242) -- (2.525,0.733);
\draw[seg,line width=0.45pt] (0.672,2.645) -- (3.299,1.008);
\draw[seg,line width=0.45pt] (1.196,3.272) -- (3.135,0.919);
\draw[seg,line width=0.45pt] (0.696,2.403) -- (2.869,0.718);
\draw[seg,line width=0.45pt] (0.892,3.215) -- (2.726,0.763);
\draw[seg,line width=0.45pt] (1.037,2.756) -- (3.080,0.976);
\draw[cstar,line width=1.1pt] (0.806,2.728) -- (3.038,1.054);
\fill[cB] (0.806,2.728) circle (2.1pt); \fill[cA] (3.038,1.054) circle (2.1pt);
\draw[xpt,line width=1.1pt,{Stealth[length=3.5pt]}-{Stealth[length=3.5pt]}] (2.698,0.828) -- (2.916,0.665);
\draw[xpt!45,line width=0.35pt,dotted] (2.929,1.136) -- (2.698,0.828);
\draw[xpt!45,line width=0.35pt,dotted] (3.147,0.972) -- (2.916,0.665);
\end{scope}
\node[anchor=east,font=\scriptsize] at (0.632,2.914) {$\bm{\mu}_2^*$};
\node[anchor=west,font=\scriptsize] at (3.410,1.271) {$\bm{\mu}_1^*$};
\node[anchor=north,font=\tiny,xpt] at (2.688,0.588) {$s_1$};
\node[font=\scriptsize,anchor=north] at (1.92,-0.34) {wide cone, short shadow};
\node[font=\scriptsize,anchor=north] at (1.92,-0.96) {$\pi$ still well determined};
\end{scope}
\begin{scope}[xshift=10.4cm]
\node[anchor=west,font=\bfseries\scriptsize] at (-0.1,3.69) {(c) spread $\parallel$ to $\bm{u}$};
\draw[gray!55,line width=0.4pt] (0,0) -- (3.91,0);
\draw[gray!55,line width=0.4pt] (0,0) -- (0,3.53);
\begin{scope}
\clip (-0.02,-0.02) rectangle (3.94,3.56);
\fill[cone,opacity=0.13] (0.806,2.728) -- ++(-42.4:7.2) arc[start angle=-42.4,end angle=-28.5,radius=7.2] -- cycle;
\draw[cB,opacity=0.55,dashed,line width=0.4pt,rotate around={-36.87:(0.806,2.728)}] (0.806,2.728) ellipse (1.178 and 0.273);
\fill[cB,opacity=0.11,rotate around={-36.87:(0.806,2.728)}] (0.806,2.728) ellipse (1.178 and 0.273);
\draw[cA,opacity=0.55,dashed,line width=0.4pt,rotate around={-36.87:(3.038,1.054)}] (3.038,1.054) ellipse (1.178 and 0.273);
\fill[cA,opacity=0.11,rotate around={-36.87:(3.038,1.054)}] (3.038,1.054) ellipse (1.178 and 0.273);
\draw[seg,line width=0.45pt] (0.320,2.867) -- (2.717,1.567);
\draw[seg,line width=0.45pt] (0.723,2.862) -- (2.992,0.793);
\draw[seg,line width=0.45pt] (1.350,2.338) -- (2.903,0.957);
\draw[seg,line width=0.45pt] (0.481,2.838) -- (2.702,1.223);
\draw[seg,line width=0.45pt] (1.293,2.642) -- (2.747,1.366);
\draw[seg,line width=0.45pt] (0.834,2.497) -- (2.960,1.012);
\draw[cstar,line width=1.1pt] (0.806,2.728) -- (3.038,1.054);
\fill[cB] (0.806,2.728) circle (2.1pt); \fill[cA] (3.038,1.054) circle (2.1pt);
\draw[xpt,line width=1.1pt,{Stealth[length=3.5pt]}-{Stealth[length=3.5pt]}] (2.336,1.100) -- (3.279,0.393);
\draw[xpt!45,line width=0.35pt,dotted] (2.567,1.407) -- (2.336,1.100);
\draw[xpt!45,line width=0.35pt,dotted] (3.509,0.701) -- (3.279,0.393);
\end{scope}
\node[anchor=east,font=\scriptsize] at (0.632,2.914) {$\bm{\mu}_2^*$};
\node[anchor=west,font=\scriptsize] at (3.410,1.271) {$\bm{\mu}_1^*$};
\node[anchor=north,font=\tiny,xpt] at (2.688,0.588) {$s_1$};
\node[font=\scriptsize,anchor=north] at (1.92,-0.34) {narrow cone, long shadow};
\node[font=\scriptsize,anchor=north] at (1.92,-0.96) {$\pi$ poorly determined};
\end{scope}
\end{tikzpicture}}
\caption{\textbf{Geometry of proportion estimation} ($K{=}2$, $G{=}2$ genes; schematic). CT profiles vary around their centroids (dashed ellipses); grey lines join each sample's two profiles, with bulk lying on the segment at position $\pi$. The orange bar is spread $s_1$ along the centroid contrast $\bm{u}$. \textbf{(a)} Tight clouds give little mixture uncertainty. \textbf{(b)} Larger variation perpendicular to the contrast changes segment orientation but leaves projected spread small. \textbf{(c)} Rotating that variation onto the contrast increases projected spread and makes proportions less well determined. The relevant separation measures compare centroid distance with residual spread, rather than total cloud size.}
\label{fig:geometry}
\end{figure}
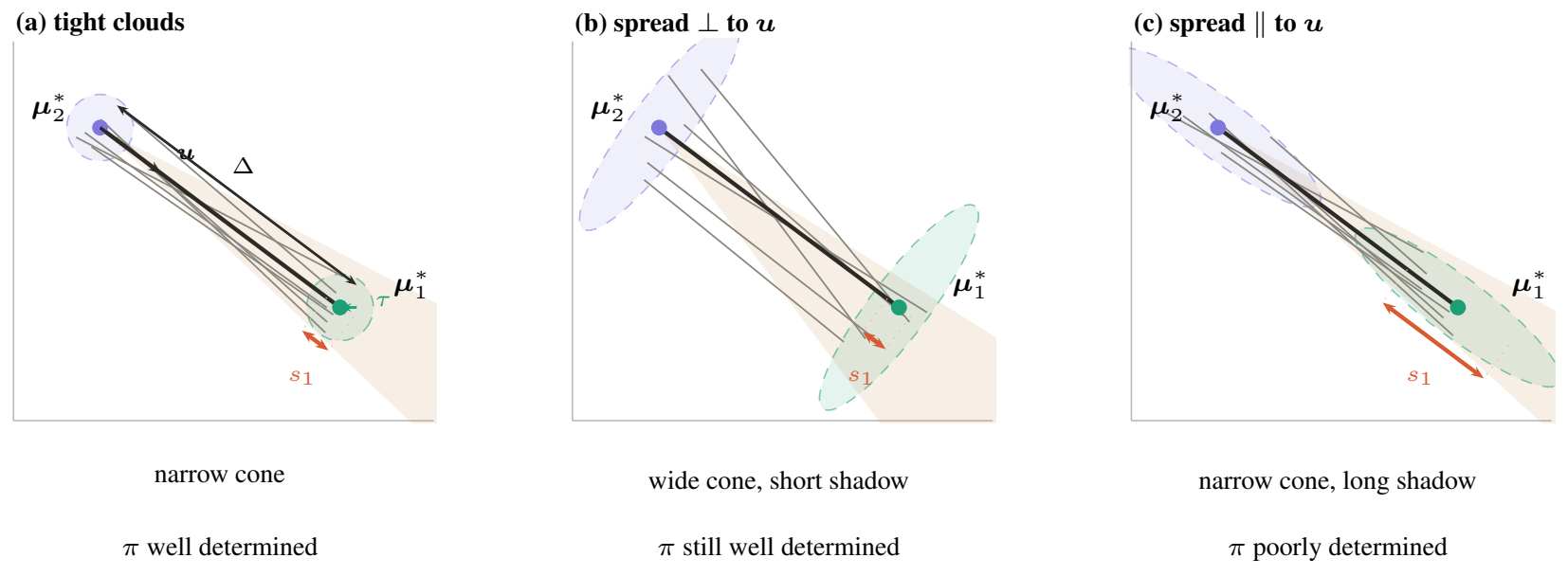

\clearpage
\begin{proof}
The steps parallel the proof of Proposition~\ref{prop:ceiling_intro} (App.~\ref{app:proof_ceiling}): compute the second moments, apply the linear-predictor formula, and read off the $R^2$. The differences are that the target is the vector $\tilde{\bm{\pi}}$ rather than a scalar, the observation is the whole profile $\bm{x}$ rather than one gene, and no Gaussian assumption is needed.

\textbf{Step 1 (moments).} By Eq.~\eqref{eq:x_contrast}, $\bm{x}-\mathbb{E}[\bm{x}]=\bm{D}(\tilde{\bm{\pi}}-\mathbb{E}[\tilde{\bm{\pi}}])+(\bm{e}-\mathbb{E}[\bm{e}])$, since the centroids and contrasts are fixed. With $\mathrm{Cov}(\bm{e},\tilde{\bm{\pi}})=\bm{0}$, the cross terms vanish:
\begin{equation}
  \mathrm{Cov}(\bm{x},\tilde{\bm{\pi}}) = \bm{D}\bm{V}, \qquad
  \mathrm{Cov}(\bm{x}) = \bm{D}\bm{V}\bm{D}^{\!\top} + \bm{T}_e .
\end{equation}

\textbf{Step 2 (best linear predictor).} Among predictors $\hat{\bm{y}} = \bm{a} + \bm{B}\bm{x}$ of a random vector $\bm{y}$, the mean squared error is minimised by $\bm{B} = \mathrm{Cov}(\bm{y},\bm{x})\,\mathrm{Cov}(\bm{x})^{-1}$ and $\bm{a} = \mathbb{E}[\bm{y}] - \bm{B}\,\mathbb{E}[\bm{x}]$ (setting the gradient to zero gives the normal equations $\bm{B}\,\mathrm{Cov}(\bm{x}) = \mathrm{Cov}(\bm{y},\bm{x})$). This is the vector form of Eq.~\eqref{eq:gauss_cond}, here as the best linear predictor, which holds for any distribution with these second moments (App.~\ref{app:varrho:floor} discusses when it is also the conditional mean). Its error $\bm{y}-\hat{\bm{y}}$ is uncorrelated with $\bm{x}$, so its covariance is
\begin{equation}
  \mathrm{Cov}(\bm{y}) - \mathrm{Cov}(\bm{y},\bm{x})\,\mathrm{Cov}(\bm{x})^{-1}\,\mathrm{Cov}(\bm{x},\bm{y}).
\end{equation}

\textbf{Step 3 (error covariance).} With $\bm{y} = \tilde{\bm{\pi}}$ and the moments of Step~1,
\begin{equation}
\label{eq:pi_err_cov_raw}
  \bm{E} = \bm{V} - \bm{V}\bm{D}^{\!\top}\bigl(\bm{D}\bm{V}\bm{D}^{\!\top} + \bm{T}_e\bigr)^{-1}\bm{D}\bm{V}.
\end{equation}
This is already exact, but it inverts a $G{\times}G$ matrix and hides the structure.

\textbf{Step 4 (Woodbury).} For invertible $\bm{A}$, $\bm{C}$ and conformable $\bm{U}$,
\begin{equation}
\label{eq:woodbury}
  (\bm{A} + \bm{U}\bm{C}\bm{U}^{\!\top})^{-1} = \bm{A}^{-1} - \bm{A}^{-1}\bm{U}\bigl(\bm{C}^{-1} + \bm{U}^{\!\top}\bm{A}^{-1}\bm{U}\bigr)^{-1}\bm{U}^{\!\top}\bm{A}^{-1},
\end{equation}
whose rank-one case is the Sherman--Morrison formula. With $\bm{A}=\bm{V}^{-1}$, $\bm{U}=\bm{D}^{\!\top}$ and $\bm{C}=\bm{T}_e^{-1}$, the right-hand side is Eq.~\eqref{eq:pi_err_cov_raw} and the left-hand side is $(\bm{V}^{-1} + \bm{D}^{\!\top}\bm{T}_e^{-1}\bm{D})^{-1}$, which is Eq.~\eqref{eq:pi_err_cov}. In this form the precision of the estimate is the prior precision $\bm{V}^{-1}$ plus the evidence $\bm{P}$ contributed by the bulk profile (cf.\ App.~\ref{app:varrho:fisher}).

\textbf{Step 5 (per-CT $R^2$).} For $k<K$, the prediction error of $\pi_k$ has variance $E_{kk}$ and $\pi_k$ has variance $V_{kk}$, so $R^2 = 1 - E_{kk}/V_{kk} = \eta_k$; because the best linear predictor is calibrated, this also equals its squared correlation with $\pi_k$. For the reference CT, $\pi_K = 1 - \bm{1}^{\!\top}\tilde{\bm{\pi}}$ is linear in $\tilde{\bm{\pi}}$, so its best linear predictor is $1 - \bm{1}^{\!\top}\hat{\tilde{\bm{\pi}}}$, with error $-\bm{1}^{\!\top}(\hat{\tilde{\bm{\pi}}} - \tilde{\bm{\pi}})$ of variance $\bm{1}^{\!\top}\bm{E}\,\bm{1}$, giving $\eta_K$.

\textbf{Reference invariance.} Choosing another reference CT replaces $\tilde{\bm{\pi}}$ by an invertible affine function of it. The best linear predictor commutes with such maps, so every $\pi_k$ receives the same prediction and error, and $\eta_k$ is unchanged.

\textbf{Two CTs and the bound on $\varrho$.} For $K{=}2$, $\bm{V}=v$ and $\bm{P}=\varrho^2$ are scalars, so $E=v/(1+v\varrho^2)$ and $\eta=1-E/v$ is Eq.~\eqref{eq:r2_pi}; moreover $E\le1/\varrho^2$ when $\varrho>0$. Cauchy--Schwarz gives $\varrho^2\ge\|\bm{\Delta}\|^2/(\bm{u}^{\!\top}\bm{T}_e\bm{u})$. With $a=\bm{u}^{\!\top}\bm{\delta}_1$ and $b=\bm{u}^{\!\top}\bm{\delta}_2$, convexity gives $(\pi a+(1-\pi)b)^2\le\pi a^2+(1-\pi)b^2\le a^2+b^2$. Hence $\bm{u}^{\!\top}\bm{T}_e\bm{u}\le s_1^2+s_2^2$, proving the stated bound without requiring composition independence.
\end{proof}
\suppressfloats[t]

\subsection{Why $\pi$ is typically easier than $\mu$}
\label{app:varrho:easier}

$\bm{P}$ pools evidence across all genes that separate the CTs. At $K{=}2$ with isotropic deviations $\bm{T}_k = \sigma_k^2\bm{I}$, $\varrho^2 = \|\bm{\Delta}\|^2/\bar\sigma^2$ with $\bar\sigma^2 := \mathbb{E}[\pi^2]\sigma_1^2 + \mathbb{E}[(1-\pi)^2]\sigma_2^2$ (for $\pi$ independent of $\bm{\delta}$ and uncorrelated $\bm{\delta}_1, \bm{\delta}_2$), which grows linearly with the number of informative genes, so $\eta \to 1$. Per-sample variation orthogonal to $\bm{u}$ does not affect $\pi$ (Fig.~\ref{fig:geometry}). By contrast, $\phi_{kg}$ is a per-gene quantity that does not improve with $G$. The advantage disappears when the contrasts are small (closely related CTs at fine resolution) or when the clouds are stretched along them (e.g.\ contrasts carried by cell-state genes); then $\pi$ is hard as well.

\subsection{Hard pairs and merging}
\label{app:varrho:merge}

\begin{figure}[tb]
\centering
\includegraphics[width=0.72\textwidth]{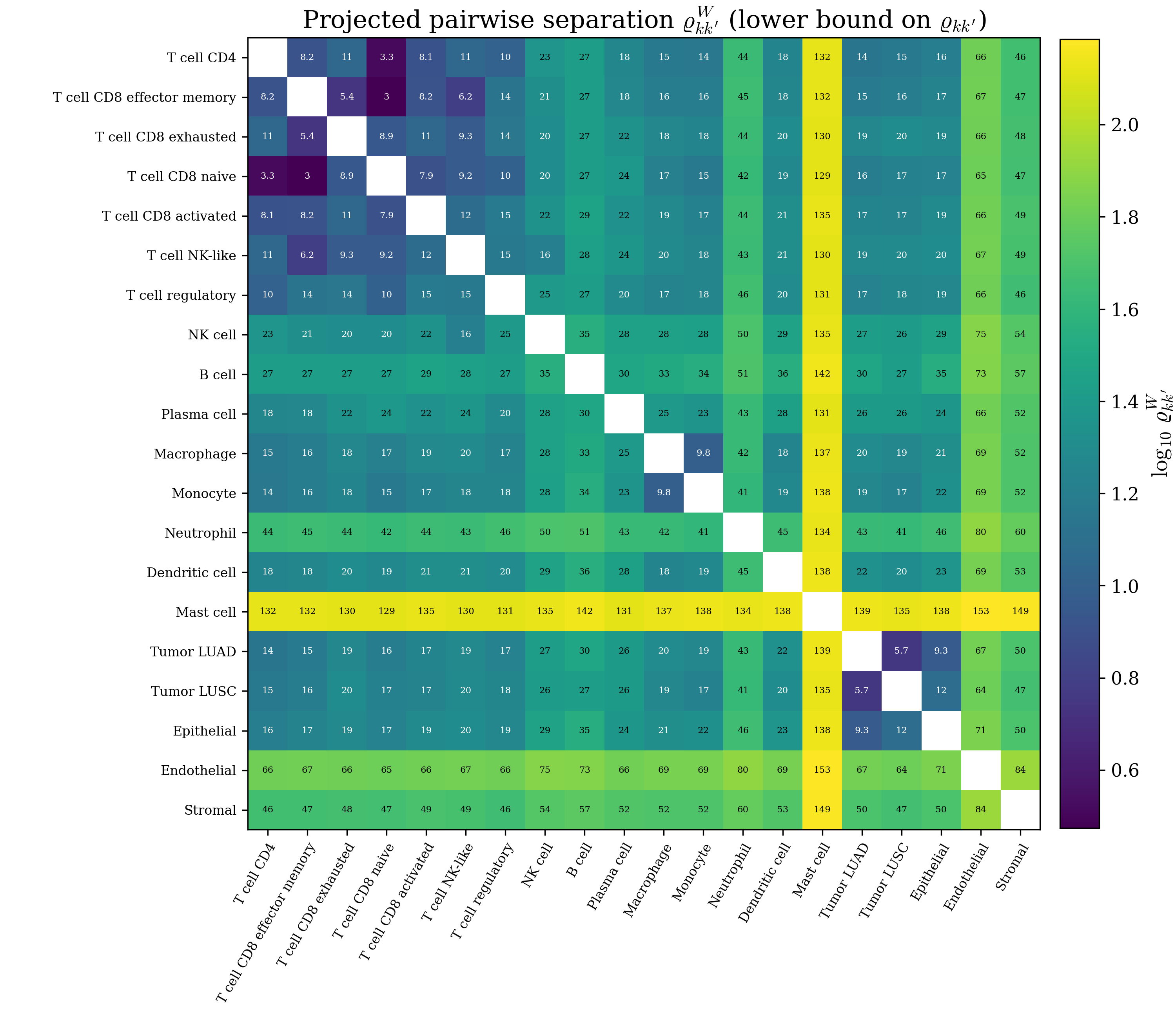}\\[2pt]
{\small (a) with whom: centroid separation $\varrho^{W}_{kk'}$}\\[6pt]
\includegraphics[width=0.72\textwidth]{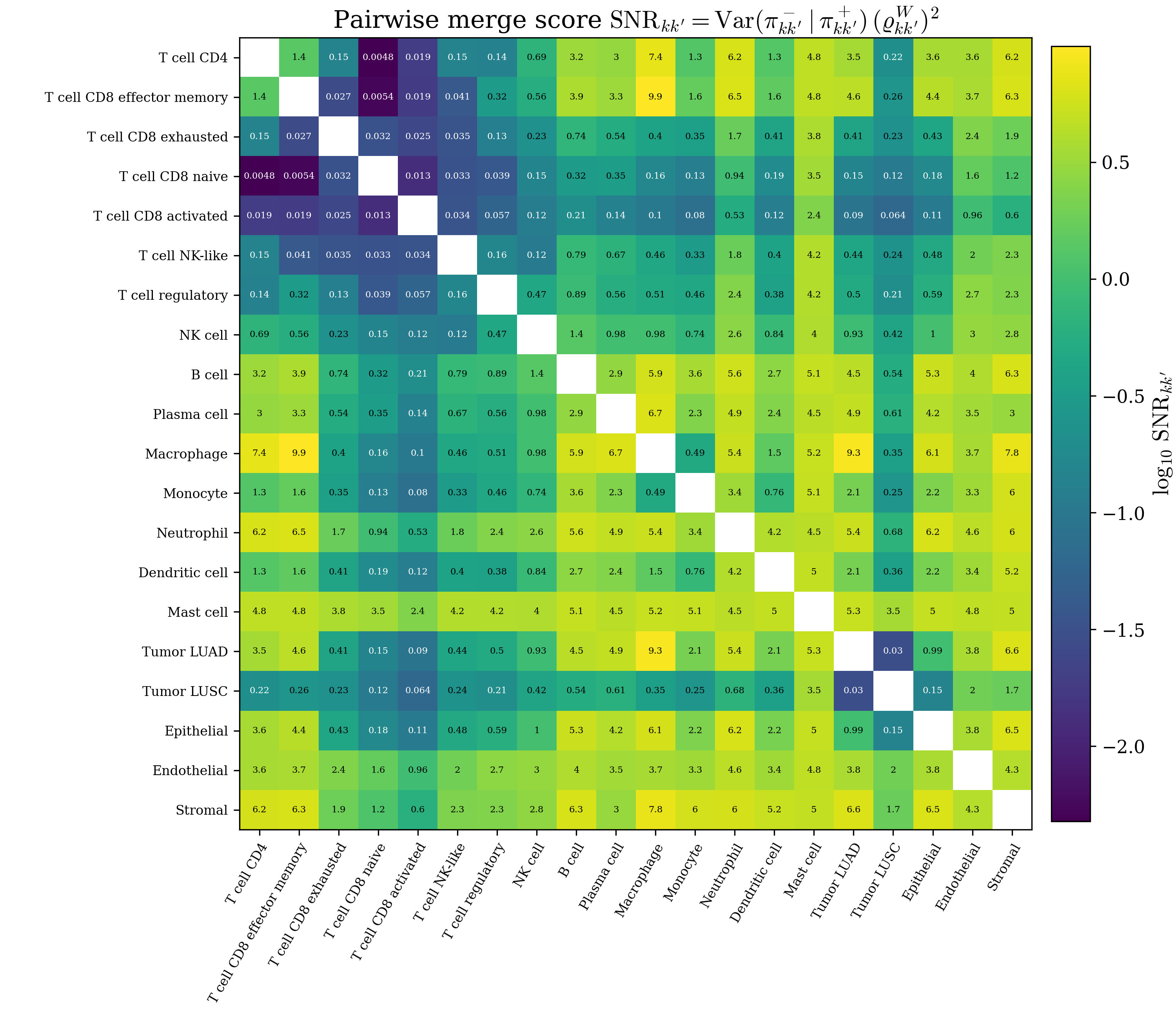}\\[2pt]
{\small (b) whether: pairwise merge score $\mathrm{SNR}_{kk'}$}
\caption{\textbf{Pairwise separation and merge score} ($K{=}20$, library-weighted; App.~\ref{app:varrho:merge}). \textbf{(a)} Projected lower bound $\varrho^{W}_{kk'}$ on $\varrho_{kk'}$; colour is $\log_{10}\varrho^{W}_{kk'}$. The least separated pairs lie among T cell subtypes and between Tumor LUAD and LUSC; mast cells are the most distinct. \textbf{(b)} $\mathrm{SNR}_{kk'}=\mathrm{Var}(\pi^{-}_{kk'}\mid\pi^{+}_{kk'})\,(\varrho^{W}_{kk'})^2$ with $\bm{V}$ from the test compositions; colour is $\log_{10}\mathrm{SNR}_{kk'}$. Rare CTs (e.g.\ CD8 activated) are low against every partner, and mast cells, though most distinct, vary little. Values are for ranking, not absolute thresholds.}
\label{fig:varrho_pairs}
\end{figure}

CT $k$ has low optimal linear accuracy when $\eta_k$ is small. Small eigenvalues of $\bm{P}$ indicate weak bulk separation in the chosen free coordinates; accuracy also depends on composition covariance $\bm{V}$. At $K=2$, $\eta\le1/2$ exactly when $v\varrho^2\le1$. For pair screening, define $\varrho_{kk'}^2=(\bm{\mu}_k^*-\bm{\mu}_{k'}^*)^{\!\top}\bm{T}_e^{-1}(\bm{\mu}_k^*-\bm{\mu}_{k'}^*)$. Under composition-independent deviations, Cauchy--Schwarz and convexity give $\varrho_{kk'}\ge\|\bm{\mu}_k^*-\bm{\mu}_{k'}^*\|/\max_j s_j(\bm{u}_{kk'})$, with $s_j(\bm{u})^2=\bm{u}^{\!\top}\bm{T}_j\bm{u}$ and $\bm{u}_{kk'}$ the unit contrast. The maximum includes all CTs, whose residual spread contributes to bulk; the two-CT bound in App.~\ref{app:varrho:main} instead uses $\sqrt{s_1^2+s_2^2}$. A small lower bound flags a pair for inspection, not proof of failure. Projecting onto $\bm{w}=\mathrm{diag}(\bm{T}_e)^{-1}(\bm{\mu}_k^*-\bm{\mu}_{k'}^*)$ gives $\varrho^W_{kk'}=|\bm{w}^{\!\top}(\bm{\mu}_k^*-\bm{\mu}_{k'}^*)|/(\bm{w}^{\!\top}\bm{T}_e\bm{w})^{1/2}\le\varrho_{kk'}$, estimated as for $\eta_k^W$. In this atlas, the projected separation is about half the diagonal approximation (median ratio $0.49$). The least separated pairs are CD8 effector memory vs CD8 naive ($3.0$), CD4 vs CD8 naive ($3.3$), CD8 effector memory vs exhausted ($5.4$), and Tumor LUAD vs LUSC ($5.7$; Fig.~\ref{fig:varrho_pairs}). Pairs confirmed difficult under the intended composition distribution are candidates for merging. Which CTs to merge follows from the same quantities. Whitening by the prior gives the signal-to-noise matrix $\bm{V}^{1/2}\bm{P}\bm{V}^{1/2}=\sum_j\lambda_j\bm{q}_j\bm{q}_j^{\!\top}$: by Eq.~\eqref{eq:pi_err_cov}, the composition contrast $\bm{q}_j^{\!\top}\bm{V}^{-1/2}\bm{\pi}$ has prior variance $1$ and error variance $1/(1+\lambda_j)$, hence optimal linear $R^2=\lambda_j/(1+\lambda_j)$. This reduces to $v\varrho^2/(1+v\varrho^2)$ at $K{=}2$, and the $\lambda_j$ do not depend on the reference CT. CTs that dominate low-$\lambda_j$ contrasts are merge candidates. A pairwise screen follows the same logic. Merging $k$ and $k'$ keeps $\pi^{+}_{kk'}:=\pi_k+\pi_{k'}$ and discards the split $\pi^{-}_{kk'}:=(\pi_k-\pi_{k'})/2$. With the other proportions held fixed, the bulk measures $\pi^{-}_{kk'}$ with precision $\varrho_{kk'}^2$, so $\mathrm{SNR}_{kk'}=\mathrm{Var}(\pi^{-}_{kk'}\mid\pi^{+}_{kk'})\,\varrho_{kk'}^2$; $\varrho_{kk'}$ alone does not say whether that precision suffices, which depends on how much the split varies across samples. The two answer different questions (Fig.~\ref{fig:varrho_pairs}): $\mathrm{SNR}_{kk'}$, like $\eta_k$, says whether a split is resolvable in a given cohort, and $\varrho_{kk'}$ says with whom to merge. For a rare CT, $\mathrm{Var}(\pi^{-}_{kk'}\mid\pi^{+}_{kk'})\approx V_{kk}$ for every more abundant partner, so its row of $\mathrm{SNR}_{kk'}$ is uniformly low while $\varrho_{kk'}$ still orders the partners. With the projected estimate of $\bm{P}$ (App.~\ref{app:varrho:eta_w}), the smallest contrasts load on CD8 naive, CD8 effector memory and CD4 T cells; on CD8 activated, CD4 and CD8 exhausted T cells; and on Tumor LUAD vs LUSC, and $\mathrm{SNR}_{kk'}$ ranks the same pairs lowest. Because the projected $\bm{P}$ is a lower bound and $\bm{V}$ depends on the cohort, we use these values for ranking rather than as absolute thresholds. Signature selection often uses CT centroids (e.g.\ signature-matrix conditioning~\cite{newman2015cibersort}); MuSiC~\cite{wang2019music}, DECALS~\cite{cai2024decals} and RNA-Sieve~\cite{erdmannpham2021rnasieve} also account for within-CT variation in fitting or inference. Proposition~\ref{prop:varrho} expresses the corresponding population linear accuracy as $R^2$.

\subsection{Relation to Fisher information}
\label{app:varrho:fisher}

For a Gaussian conditional model with composition-independent residual covariance, $\bm{P}$ is the Fisher information for $\tilde{\bm{\pi}}$; when invertible, $\bm{P}^{-1}$ gives the Cram\'er--Rao lower bound for unbiased estimation. The linear prediction error $\bm{E}=(\bm{V}^{-1}+\bm{P})^{-1}$ resembles Gaussian prior precision updating. This is not a general van Trees bound with prior information $\bm{V}^{-1}$ for a simplex-supported distribution. Composition-dependent conditional covariance contributes an additional Fisher-information term, which linear predictors do not explicitly exploit.

\subsection{No distributional assumption}
\label{app:varrho:nongauss}

Proposition~\ref{prop:varrho} uses second moments only. This matters because per-CT clouds cannot be close to Gaussian at every CT resolution at once: merging sub-types into a parent CT turns sample-to-sample variation in the sub-type composition into variation of the parent's expression profile, which is generally a non-Gaussian mixture. 
Linear predictors are a subset of all predictors, so the Bayes-optimal MSE is at most the best linear MSE, with equality precisely when $\mathbb{E}[\tilde{\bm{\pi}}\mid\bm{x}]$ is affine in $\bm{x}$~\cite{kay1993estimation}. Non-degenerate simplex-supported proportions are not Gaussian, but non-Gaussianity alone does not guarantee a nonlinear gain. With composition-independent deviations, the conditional covariance $\mathrm{Cov}(\bm{e}\mid\bm{\pi})=\sum_{k,k'}\pi_k\pi_{k'}\mathrm{Cov}(\bm{\delta}_k,\bm{\delta}_{k'})$ can vary with composition, providing structure beyond the linear predictor's second-moment calculation.

\subsection{Ceiling or floor?}
\label{app:varrho:floor}

$\phi^{\mathrm{CT}}_{kg}$ at fixed composition and $\eta_k$ are optimal linear $R^2$ values in different information classes: one gene plus composition for CTSE, and the whole profile for proportions. The independent-CT expression case is $\phi_{kg}$. These linear results require second moments, not Gaussianity. Each is a ceiling for its linear predictor class and a lower bound on optimal nonlinear performance in the same information class. If the conditional mean is affine, as under joint Gaussianity, the linear optimum also bounds nonlinear predictors. None is a guaranteed floor for an arbitrary fitted model.

In our atlas, per-sample CT means are strongly non-Gaussian in linear CPM: median skewness $3.1$ and excess kurtosis $15$ over 106 targets; the Hartigan dip test rejects unimodality for 26\%. A finite-sample comparison of gradient boosting on $(x_g,\bm{\pi})$ with the atlas-based Gaussian-form predictor gives median held-out $R^2$ improvement $0.007$ (IQR $-0.091$ to $0.084$). The gain is unrelated to kurtosis. A sample-group bootstrap 95\% interval excludes zero in favour of boosting for 16 pairs (including stromal COL1A1 and POSTN and CD8 effector-memory GZMB), and in favour of the Gaussian-form predictor for 8. This comparison does not establish conditional linearity or a nonlinear ceiling: fitted predictors, atlas parameter error, finite-cell variation and split differences can all affect performance.
For proportions, $\eta_k$ remains the linear optimum regardless of distribution. Ridge regression on linear-CPM inputs belongs to the predictor class of Proposition~\ref{prop:varrho}; its accuracy tracks $\eta_k^{W}$ ($\rho{=}0.81$) and matches the MLP on average (median $R^2$ $0.85$ vs $0.82$). Ridge penalties are selected per CT by five-fold CV grouped by atlas sample, with input standardisation fitted on training samples.

Both results guide screening: a large score indicates that accurate linear prediction is possible with population parameters, while a small score limits only the specified linear class unless the stronger conditional-mean assumptions hold. Finite-sample plug-in estimates need not equal these population quantities.

\subsection{When composition and expression co-vary}
\label{app:varrho:covary}

If $\mathrm{Cov}(\bm{e},\tilde{\bm{\pi}})\neq\bm{0}$, decompose $\bm{e}=\bm{\Gamma}(\tilde{\bm{\pi}}-\bar{\tilde{\bm{\pi}}})+\bm{e}_\perp$, where $\bm{\Gamma}=\mathrm{Cov}(\bm{e},\tilde{\bm{\pi}})\bm{V}^{-1}$ and $\mathrm{Cov}(\bm{e}_\perp,\tilde{\bm{\pi}})=\bm{0}$. Proposition~\ref{prop:varrho} then holds with $\bm{D}\to\bm{D}+\bm{\Gamma}$ and $\bm{T}_e\to\bm{T}_e-\bm{\Gamma}\bm{V}\bm{\Gamma}^{\!\top}$. Coupled expression shifts can strengthen or weaken the effective contrast. Such coupling is expected in tissue (e.g.\ more T cells together with interferon-driven states), but resampling may not preserve its biological distribution~\cite{hu2024heterogeneous}. Dirichlet resampling alone does not guarantee $\bm{\Gamma}=0$, particularly with sample-specific composition means and library-weighted fractions. If coupling changes between pseudobulk and bulk, a fixed linear predictor's response to composition changes through $\bm{W}(\bm{D}+\bm{\Gamma})$, potentially distorting estimates. Composition shift can also induce centroid shift when $\mathbb{E}[\bm{\delta}_k\mid\bm{\pi}]$ varies with composition.

\subsection{A computable lower bound}
\label{app:varrho:eta_w}

\begin{figure}[tb]
\centering
\includegraphics[width=\textwidth]{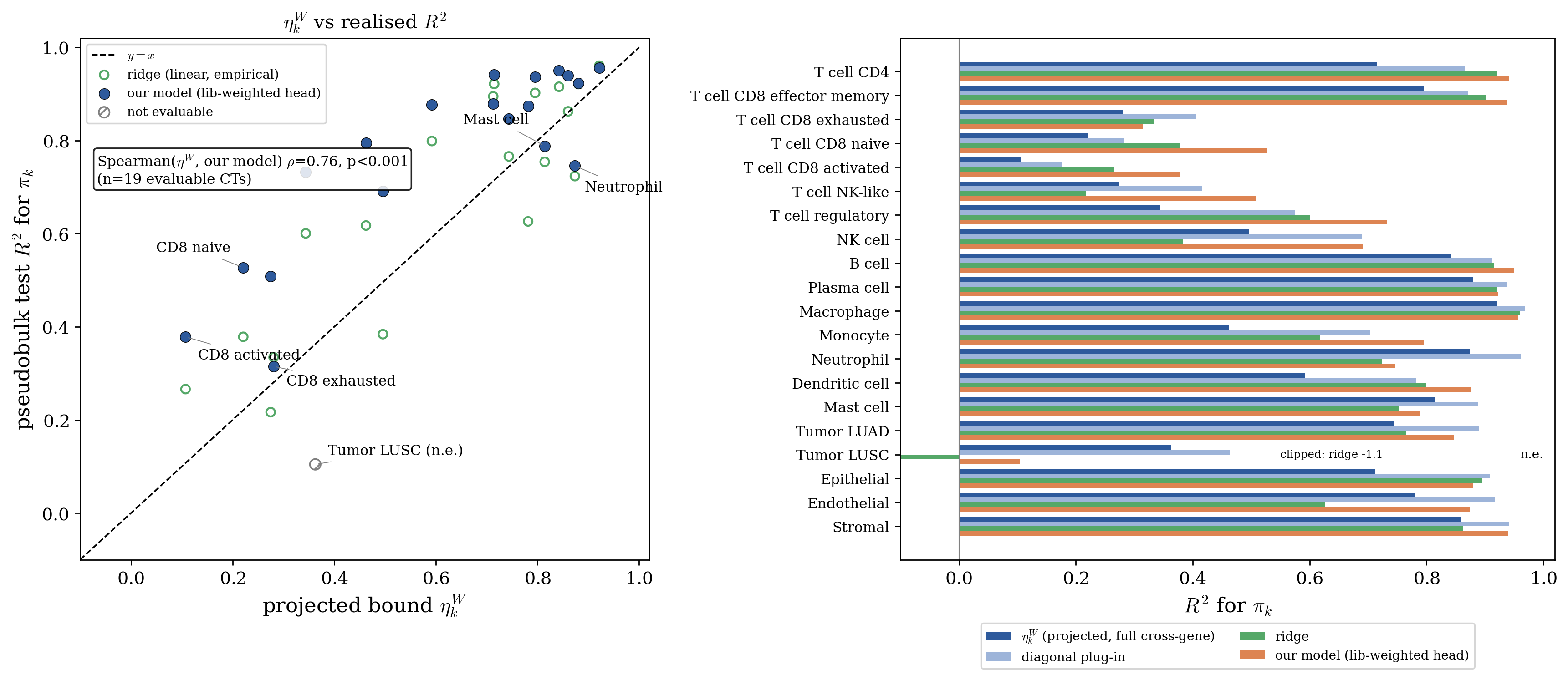}
\caption{\textbf{Projected proportion score vs observed accuracy} ($K{=}20$, library-weighted proportions). \emph{Left}: plug-in $\eta_k^{W}$ against per-CT test $R^2$ of our model (filled) and ridge (open). The population lower bound uses $K{-}1$ linear-CPM projections; the plotted ridge uses log-CPM inputs, so its predictor class differs (ridge on linear-CPM inputs, discussed in the text, reaches median $R^2$ $0.85$). LUSC is excluded from the correlation because its test/train proportion-variance ratio is below $0.1$: only one of 35 test samples has mean LUSC fraction above 5\%. Its small test variance makes $R^2$ sensitive to bias (ridge $-1.06$, MLP $0.11$). \emph{Right}: the projected estimate, the diagonal approximation (larger in this atlas), ridge and MLP accuracy.}
\label{fig:eta}
\end{figure}

Equation~\eqref{eq:pi_err_cov} requires unregularized inversion of the $G{\times}G$ residual covariance, which is underdetermined at this atlas size. Instead, fix weights $\bm{W}\in\mathbb{R}^{(K-1)\times G}$ and predict $\tilde{\bm{\pi}}$ from $\bm{z}=\bm{W}\bm{x}$. The projected covariance is $\mathrm{Cov}(\bm{z})=\bm{W}\bm{D}\bm{V}\bm{D}^{\!\top}\bm{W}^{\!\top}+\bm{W}\bm{T}_e\bm{W}^{\!\top}$ and $\mathrm{Cov}(\bm{z},\tilde{\bm{\pi}})=\bm{W}\bm{D}\bm{V}$, giving $\bm{E}_W=\bm{V}-\bm{V}\bm{D}^{\!\top}\bm{W}^{\!\top}\mathrm{Cov}(\bm{z})^{-1}\bm{W}\bm{D}\bm{V}$. Thus $\eta_k^{W}=1-[\bm{E}_W]_{kk}/V_{kk}\le\eta_k$ for $k<K$; for the reference CT use $1-\bm{1}^{\!\top}\bm{E}_W\bm{1}/\mathrm{Var}(\pi_K)$. This population lower bound is attainable by a predictor of the projected scores. We estimate it from atlas sample means and projected cells, using $\bm{W}=\bm{D}^{\!\top}\mathrm{diag}(\bm{T}_e)^{-1}$ on 1,024 genes, finite-cell sampling variance in $\bm{T}_e$, and test-split $\bm{V}$. The plug-in median is $0.71$; the MLP exceeds it for 17 of 20 CTs, with $\rho=0.76$ ($p=2\times10^{-4}$) over 19 evaluable CTs. Neutrophil and Mast cell fall below the estimate; LUSC is excluded by the variance rule in Fig.~\ref{fig:eta}. Empirical departures can reflect estimation error and train/test distribution differences rather than a violation of the population bound.

\subsection{Oracle assumptions and error propagation}
\label{app:error_propagation}

Both results assume oracle inputs, and that all relevant populations are represented in the reference; missing types can make targets non-identifiable even without sampling error~\cite{jiang2026incomplete}. The expression ceiling (Prop.~\ref{prop:ceiling_intro}) conditions on the true centroids $\mu_{kg}^*$, the sample's composition $\bm{\pi}^{(i)}$ and the population covariance $\bm{\Sigma}_g$; the proportion result (Prop.~\ref{prop:varrho}) assumes known contrasts $\bm{D}$, residual covariance $\bm{T}_e$ and composition covariance $\bm{V}$. For an estimator that sees only the bulk $\bm{x}$, the true $\bm{\pi}^{(i)}$ is extra information, so in this respect the expression ceiling is optimistic. In practice all these inputs are estimated, and their errors propagate.

\textbf{Expression.} Errors in centroids, covariance and composition affect both the residual and the shrinkage coefficient of the oracle predictor. A centroid error enters directly and through the subtraction $x_g-\sum_j\pi_j\mu^*_{jg}$; composition errors alter both that subtraction and the coefficient. These contributions can be correlated, so the oracle error is not an additive error budget for fitted estimators. Rare CTs can have large relative composition errors, but absolute sensitivity depends on expression contrasts and shrinkage. In particular, the diagonal coefficient $\phi_{kg}/\pi_k=\pi_k\tau_{kg}^2/\sum_j\pi_j^2\tau_{jg}^2$ can shrink as $\pi_k$ decreases, rather than generally diverging as $1/\pi_k$. Likewise, low $\eta_k$ indicates relative prediction difficulty: the optimal linear absolute error variance is $(1-\eta_k)\mathrm{Var}(\pi_k)$.

\textbf{Proportions.} For a fixed linear predictor $\bm{W}$, contrast shift $\Delta\bm{D}$ contributes $\bm{W}\Delta\bm{D}\,\tilde{\bm{\pi}}$ to prediction shift, possibly distorting scale while preserving ranking. Changes in the reference centroid also shift the intercept. Covariance estimation errors make fitted predictors suboptimal, while finite-sample estimates of $\eta_k$ and $\phi^{\mathrm{CT}}_{kg}$ have their own uncertainty. Neither the sign nor size of plug-in bias is guaranteed. Composition--state coupling adds a further source of shift (App.~\ref{app:varrho:covary}).

\FloatBarrier
\section{Prognostic Analyses}
\label{app:cox}

\subsection{Top Prognostic Features}
\label{app:features}

\begin{figure}[tb]
\centering
\includegraphics[width=0.9\textwidth]{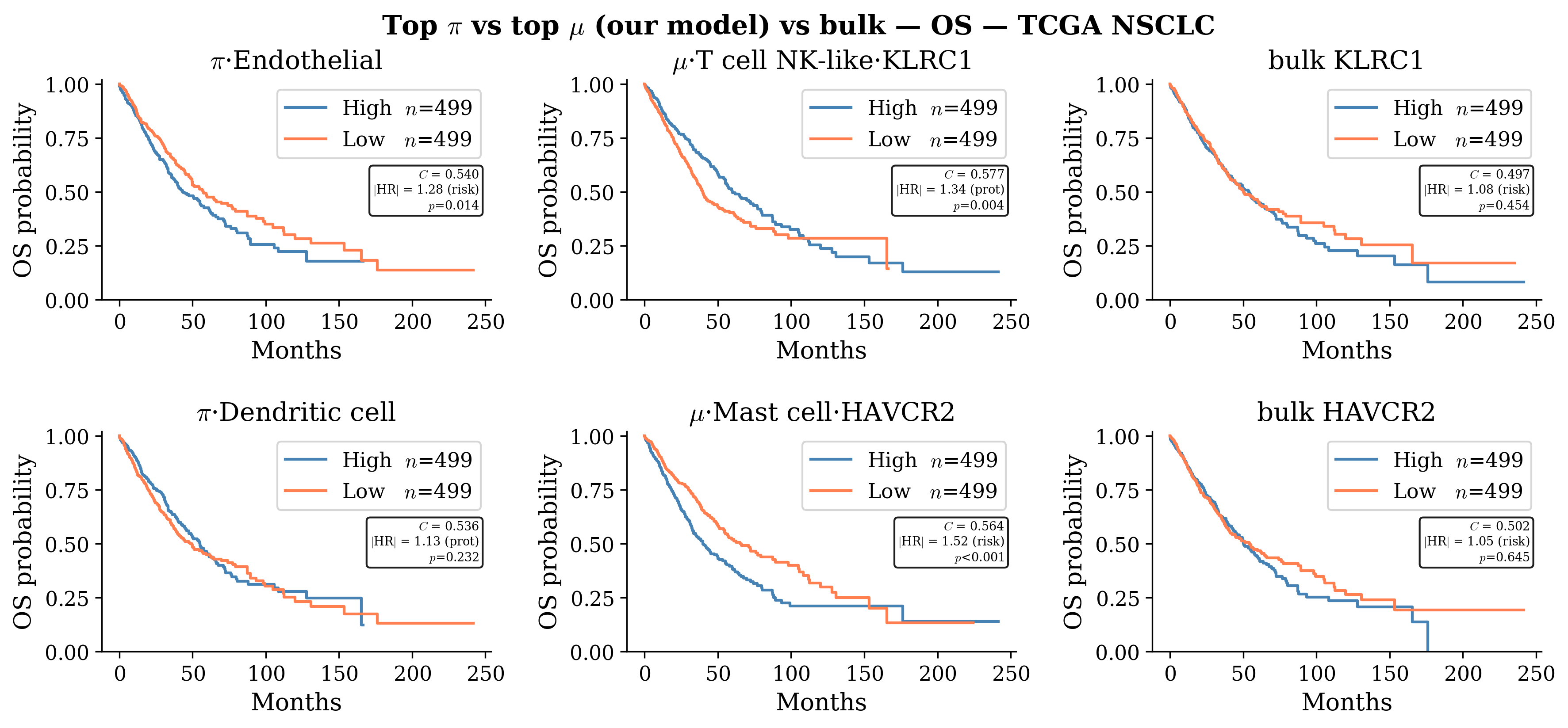}
\caption{\textbf{Kaplan--Meier curves for top single features (selected post hoc, primary model).} Patients stratified by median split; nominal log-rank $p$-values. Rows: $\hat\pi$ Endothelial, $\hat\mu$ KLRC1 in NK-like T cells, bulk KLRC1 ($p{=}0.014$, $0.0039$, $0.45$); $\hat\pi$ Dendritic cell, $\hat\mu$ HAVCR2 in mast cells, bulk HAVCR2 ($p{=}0.23$, $3.6{\times}10^{-5}$, $0.65$).}
\label{fig:km_top}
\end{figure}

All analyses use TCGA NSCLC OS ($n{=}998$, 394 events) and our primary model unless stated; features were selected post hoc from the 106 $\hat\mu_{kg}$ and 20 $\hat\pi_k$ (Table~\ref{tab:top_features}, Fig.~\ref{fig:km_top}).

\textbf{HAVCR2 (TIM-3) in mast cells} (HR $1.22$ per SD vs $1.04$ for bulk HAVCR2): high predicted mast-cell HAVCR2 is associated with worse survival. Mast cells can promote progression via pro-angiogenic factors, and TIM-3$^{+}$ mast cells may represent an activated state.

\textbf{KLRC1 (NKG2A) in NK-like T cells} (HR $0.79$ per SD vs $0.97$ for bulk KLRC1): high NK-like T cell KLRC1 is associated with better OS. KLRC1/NKG2A is an inhibitory checkpoint, making this association counterintuitive; we report it as an empirical finding. It is the most consistent $\hat\mu_{kg}$ feature across models (App.~\ref{app:controls}).

\textbf{Other features.} The next-best single $\hat\mu_{kg}$ features are CXCL13 in CD4 T cells ($C$ $0.555$ vs bulk $0.542$), CDH1 in Tumor LUAD ($0.555$ vs $0.525$) and PRF1 in CD8 effector memory T cells ($0.546$ vs $0.511$). The best $\hat\pi_k$ features are Endothelial ($0.540$), Dendritic cell ($0.536$) and Monocyte ($0.536$).

\begin{table}[h]
\centering
\small
\caption{Top single features (post hoc, primary model). $C$: 5-fold out-of-fold univariate Cox; HR per SD; $q$: Benjamini--Hochberg within family (106 $\mu$, 20 $\pi$). Robust: meets the across-model criterion in App.~\ref{app:controls}, which uses nominal significance; other robust features (e.g.\ IL2RA in Tregs) are listed there. Only KLRC1 in NK-like T cells is both robust and significant after BH correction.}
\label{tab:top_features}
\begin{tabular}{llcccccc}
\toprule
Gene & Cell type & $C_{\text{bulk}}$ & $C_{\text{deconv}}$ & HR$_{\text{bulk}}$ & HR$_{\text{deconv}}$ & $q$ & Robust \\
\midrule
\multicolumn{8}{l}{\textit{$\hat\mu_{kg}$ features}} \\
KLRC1 & T cell NK-like & 0.497 & 0.577 & 0.97 & 0.79 & $3{\times}10^{-4}$ & \checkmark \\
HAVCR2 & Mast cell & 0.502 & 0.564 & 1.04 & 1.22 & 0.010 & \xmark \\
CXCL13 & T cell CD4 & 0.542 & 0.555 & 0.87 & 0.84 & 0.014 & \xmark \\
CDH1 & Tumor LUAD & 0.525 & 0.555 & 0.94 & 0.87 & 0.14 & \xmark \\
PRF1 & T cell CD8 EM & 0.511 & 0.546 & 1.08 & 1.14 & 0.17 & \checkmark \\
\midrule
\multicolumn{8}{l}{\textit{$\hat\pi_k$ features}} \\
--- & Endothelial & --- & 0.540 & --- & 1.13 & 0.090 & \checkmark \\
--- & Dendritic cell & --- & 0.536 & --- & 0.90 & 0.17 & \xmark \\
--- & Monocyte & --- & 0.536 & --- & 1.11 & 0.16 & \checkmark \\
\bottomrule
\end{tabular}
\end{table}

\subsection{Controls and Model Robustness}
\label{app:controls}

Because $\hat\mu_{kg}$ is a function of bulk and random bulk-derived signatures can be prognostic~\cite{venet2011random}, we assess adjustment and model robustness. These are internal checks on the same TCGA cohort, not independent clinical replication.

\textbf{Adjustment.} Cox models include $\hat\mu_{kg}$, $\hat\pi_k$, matched bulk $x_g$ and proliferation (mean of z-scored MKI67, TOP2A, CCNB1, BUB1, CDK1, CCNA2, AURKA, MCM2, PCNA, TYMS: ten canonical cell-cycle genes; in breast cancer, proliferation drives much of the prognostic signal of random signatures~\cite{venet2011random}; the adjustment is a sensitivity analysis), using ensemble predictions. KLRC1 in NK-like T cells has a nominal unadjusted association (HR $0.84$, 95\% CI $0.76{-}0.93$, $p=6\times10^{-4}$); after adjustment it survives BH correction (HR $0.76$, $0.67{-}0.85$, $p=5\times10^{-6}$, $q=5\times10^{-4}$ across 106). No unadjusted ensemble CTSE association passes BH. Most associations that emerge after adjustment are suppression effects: their coefficients contrast CTSE predictions with correlated bulk expression. Further adjusting for age, sex, pathological stage and histology (stratified baseline hazards; complete cases, $n=970$, 387 events) gives HR $0.79$ ($0.69{-}0.89$) for KLRC1, versus $0.96$ ($0.85{-}1.08$) for bulk KLRC1. The direction holds in LUAD ($0.85$, $0.70{-}1.03$) and LUSC ($0.80$, $0.67{-}0.94$) separately, and proportional hazards are not rejected (Schoenfeld $p=0.79$).

\textbf{Robustness across models.} We select checkpoints within 10\% of the best pseudobulk median test $R^2$ for both CTSE and library-weighted proportions, restricted to matching inputs and normalisation: three checkpoints spanning two architectures and loss weightings. The ensemble averages their TCGA predictions. A feature meets our robustness criterion if its direction agrees in all three, it is nominally significant ($p<0.05$) in at least two and in the ensemble, and, for CTSE, its ensemble $C$ and $|\log\mathrm{HR}|$ exceed those of the matched bulk gene. Seven features qualify: CTSE KLRC1 in NK-like T cells (ensemble $C=0.559$, HR $0.84$), NCR1 in NK-like T cells ($0.543$, $1.11$), PRF1 in CD8 effector memory T cells ($0.542$, $1.15$), and IL2RA in Tregs ($0.534$, $1.11$); proportions Monocyte ($0.540$, $1.11$), Endothelial ($0.531$, $1.14$), and NK cell ($0.512$, $1.13$). The qualifying set changes if a fourth, near-threshold checkpoint is included. KLRC1 is selected in 94\% of internal LASSO-Cox fits (25 training folds $\times$ 10 bootstraps), the only feature above 70\%; those overlapping fits are not independent replications.

\textbf{Sparse model comparison.} Repeated nested CV gives mean $C=0.584$ for ensemble CTSE plus proportions, $0.582$ for CTSE alone, $0.564$ for proportions alone, and $0.570$ for bulk LASSO on the model input genes. The deconvolved and bulk models select median 15 and 18 features, respectively: candidate representations differ, but both fitted models are sparse. Adding deconvolved features to bulk gives $C=0.572$, little change from bulk alone. Patient-level bootstrap intervals for paired differences: CTSE plus proportions vs the 1{,}024 bulk genes $+0.014$ ($-0.010$, $0.040$); vs the 55 raw target genes plus proportions ($C=0.553$) $+0.031$ ($0.008$, $0.052$). With age, sex, stage and histology as unpenalised covariates ($n=970$), $C$ is $0.629$, $0.620$ and $0.633$ for these three models, against $0.617$ for the clinical covariates alone: CTSE plus proportions still exceeds the matched baseline ($+0.010$, $0.001$ to $0.018$) but not the 1{,}024-gene model ($-0.004$, $-0.016$ to $0.009$), and no molecular model adds more than $0.016$ to the clinical covariates. Deconvolution still consumes the full bulk input profile, so downstream sparsity does not imply a smaller gene assay.

\FloatBarrier
\section{Robustness Checks}
\label{app:robust}

\subsection{Donor-held-out replicate}
\label{app:robust:donor}
We re-split the 512 atlas samples by donor (378 training, 44 validation and 90 test donors; study-stratified, whole donors per partition) and refit every data-driven step on training donors only: input-gene ranking, feature normalisation, the scores $\phi_{kg}$, $\phi^{\mathrm{CT}}_{kg}$ and $\eta_k^{W}$, and checkpoint selection on validation donors. The configuration is unchanged. Test donors are evaluated once, with donor-level bootstrap intervals (Table~\ref{tab:donor}). Only three test donors are LUSC-dominant, and this is a single split and seed. The ranked part of the input-gene list is split-dependent (420 of 1{,}024 genes shared with the main model), while the conclusions are not.

\begin{table}[h]
\centering
\small
\caption{Donor-held-out replicate vs the main (sample-split) model. Medians over 106 targets or 20 CTs; Spearman $\rho$ with per-target (CTSE) or per-CT (proportion) test $R^2$.}
\label{tab:donor}
\begin{tabular}{lcc}
\toprule
 & donor-held-out (95\% CI) & main \\
\midrule
CTSE $R^2$, $\log_2$ & $0.197$ ($0.146$, $0.226$) & $0.19$ \\
CTSE $R^2$, linear CPM: raw / calibrated on validation donors & $0.006$ / $0.139$ & $-0.01$ / -- \\
proportion $R^2$, model & $0.760$ ($0.741$, $0.824$) & $0.82$ \\
proportion $R^2$, linear-CPM ridge & $0.831$ ($0.781$, $0.879$) & $0.85$ \\
$\rho(\phi_{kg},\,R^2)$ & $0.63$ ($0.50$, $0.66$) & $0.49$ \\
$\rho(\phi^{\mathrm{CT}}_{kg},\,R^2)$ & $0.44$ ($0.33$, $0.52$) & $0.39$ \\
$\rho(\eta_k^{W},\,R^2)$, 19 evaluable CTs & $0.81$ ($0.64$, $0.89$) & $0.76$ \\
\bottomrule
\end{tabular}
\end{table}

\subsection{Learnability scores, simple proxies and data requirements}
\label{app:robust:proxies}
On the donor-held-out split we compared the scores with simple atlas quantities, all estimated on training donors: mean CT abundance $\bar\pi_k$; the target CT's mean-expression share, $\xi_{kg} := \bar\pi_k\mu^*_{kg}/\sum_j \bar\pi_j\mu^*_{jg}$; and the gene's variance. Their Spearman correlations with per-target $\log_2$ test $R^2$ are $0.50$, $0.60$ and $0.39$, against $0.63$ for $\phi_{kg}$ and $0.44$ for $\phi^{\mathrm{CT}}_{kg}$; given the three simple quantities, the partial correlation of $\phi_{kg}$ is $0.15$ ($-0.03$, $0.24$) and of $\phi^{\mathrm{CT}}_{kg}$ $-0.07$. The share $\xi_{kg}$ is the mean-expression analogue of $\phi_{kg}=\pi_k^2\tau_{kg}^2/\sum_j\pi_j^2\tau_{jg}^2$, which explains their agreement; $\phi_{kg}$, unlike $\xi_{kg}$, has an accuracy interpretation and accounts for how much the target CT's expression actually varies. The three differ in what they require. $\xi_{kg}$ needs only CT mean profiles, available from any reference or signature matrix; $\phi_{kg}$ needs between-sample variances of CT means, hence several samples per CT; $\phi^{\mathrm{CT}}_{kg}$ needs a full $K{\times}K$ covariance per gene (210 parameters at $K{=}20$) and is the noisiest with few donors.

\subsection{Per-gene linear estimators against the ceiling}
\label{app:robust:linear}
To test Proposition~\ref{prop:ceiling_intro} in its own estimator class, we fitted the TCA form (diagonal $\bm{\Sigma}_g$) and the Unico form (full $\bm{\Sigma}_g$) of the conditional-mean estimator, with centroids, covariances and finite-cell variance from training donors, and evaluated them in linear CPM on test donors (Table~\ref{tab:linear}). Their accuracy tracks $\phi^{\mathrm{CT}}_{kg}$ and stays below it for most targets; the few exceedances are consistent with estimation error in a population quantity and with test-sample fluctuation. They exceed the diagonal $\phi_{kg}$ more often, as expected when CTs are correlated, since $\phi_{kg}$ is the ceiling only for independent CTs. They also beat the calibrated network in linear space for 51 (TCA) and 72 (Unico) of 106 targets.

\begin{table}[h]
\centering
\small
\caption{Per-gene linear estimators on donor-held-out test samples (linear CPM; median $\phi^{\mathrm{CT}}_{kg}=0.42$, median $\phi_{kg}=0.06$ over the 106 targets).}
\label{tab:linear}
\begin{tabular}{llccc}
\toprule
estimator & composition & median $R^2$ & $\rho$ with $\phi^{\mathrm{CT}}_{kg}$ & targets with $R^2>\phi^{\mathrm{CT}}_{kg}$ \\
\midrule
TCA form & true & $0.109$ & $0.58$ & 9/106 \\
Unico form & true & $0.205$ & $0.61$ & 14/106 \\
TCA form & predicted & $0.079$ & $0.47$ & 8/106 \\
Unico form & predicted & $0.157$ & $0.46$ & 15/106 \\
\bottomrule
\end{tabular}
\end{table}

\subsection{Linear baselines vs the network}
\label{app:robust:c2}
On the same donor-held-out test samples we compared the network with the per-gene linear estimators of App.~\ref{app:robust:linear} (fed with the network's predicted proportions) and with a per-target ridge regression on the same 1{,}024 $\log_2$ inputs (penalty by donor-grouped CV on training donors). Table~\ref{tab:c2} gives median per-target $R^2$ in both spaces. The ranking reverses between spaces: the additive linear estimators are best in linear CPM but poor in $\log_2$ space, where small positive means are hard to represent, whereas the network is poorly calibrated in linear CPM. The log-input ridge beats the network in $\log_2$ space (median difference $+0.087$, $0.067{-}0.111$; ahead for 90 of 106 targets), so the network is the most accurate CTSE predictor in neither space, as for proportions (App.~\ref{app:robust:donor}). Accuracy patterns nonetheless agree across models (Spearman $0.85$ between network and ridge per-target $R^2$), and the scores track the ridge's accuracy as well as the network's ($\rho=0.65$ for $\phi_{kg}$, $0.51$ for $\phi^{\mathrm{CT}}_{kg}$, $0.62$ for $\xi_{kg}$): per-target difficulty reflects learnability rather than the choice of model.

\begin{table}[h]
\centering
\small
\caption{Median per-target CTSE $R^2$ on donor-held-out test samples (95\% donor-bootstrap CI).}
\label{tab:c2}
\begin{tabular}{lcc}
\toprule
method & $\log_2(\mathrm{CPM}{+}1)$ & linear CPM \\
\midrule
network & $0.197$ ($0.146$, $0.226$) & $0.006$ ($-0.025$, $0.038$) \\
TCA form, predicted composition & $-0.247$ ($-0.319$, $-0.199$) & $0.079$ ($0.043$, $0.095$) \\
Unico form, predicted composition & $-0.157$ ($-0.228$, $-0.131$) & $0.157$ ($0.097$, $0.170$) \\
per-target ridge, $\log_2$ inputs & $0.280$ ($0.251$, $0.310$) & $-0.016$ ($-0.057$, $0.012$) \\
\bottomrule
\end{tabular}
\end{table}

\subsection{Configuration sweep}
\label{app:robust:sweep}
Six one-factor changes to the configuration (proportions as an input to the CTSE heads, trunk dropout $0.2$, proportion-head dropout $0.3$, a doubled expression or proportion loss weight, and trunk width 1{,}024) were compared on validation donors by median CTSE $R^2$. None improved on the baseline ($0.161$; alternatives $0.123{-}0.158$), so the main configuration is kept. Doubling the proportion loss weight raised validation proportion $R^2$ ($0.91$ vs $0.82$) at a small CTSE cost; it was not evaluated on test donors.

\FloatBarrier
\section{Detailed Methods}
\label{app:methods}

\subsection{Pseudobulk Training Data}
We generated pseudobulk samples from a NSCLC single-cell atlas~\cite{salcher2022nsclc} (${\sim}1.28$M cells, 29 datasets; the 512 samples with ${\ge}100$ cells, from 307 donors) via Dirichlet sampling with concentration $\alpha{=}30$. Each pseudobulk sample $i$ has known cell type proportions $\bm{\pi}^{(i)} \in \Delta_{K-1}$ ($K{=}20$ cell types) and per-cell-type gene expression $\mu_{kg}^{(i)}$. Bulk expression is the exact mixture:
\[
x_g^{(i)} = \sum_{k=1}^K \pi_k^{(i)} \mu_{kg}^{(i)}
\]
in linear counts-per-million (CPM) space. 
We transform to $\log_2(\text{CPM}+1)$ for training due to better numerical stability. 
Splits are by atlas sample: about 45k pseudobulk samples from 460 training samples, 1{,}365 pseudobulk samples from 35 held-out test samples, and a further 17 held-out samples used for validation (early stopping and checkpoint selection). Of these 35, 28 share a donor with training (e.g.\ tumour and adjacent normal), so test $R^2$ may be optimistic for new donors. TCGA is an independent cohort, although this overlap limits the validation of estimation accuracy. Empty sample--CT groups and unselected pairs are masked from expression training and evaluation.
Inputs: 1{,}024 genes, namely the 56 target genes, 47 curated lineage markers, and the top-ranked remaining genes by XGBoost importance for the proportions, computed on training samples only.  

\subsection{Model Architecture and Target Selection}
Skip-connection multilayer perceptron (MLP) with separate heads for proportions $\hat{\bm{\pi}}$ (sigmoid followed by renormalisation onto the simplex; cell-count and library-weighted versions, the latter used throughout), average expression $\hat{\bm{\mu}}_{kg}$ per (CT, gene) pair, and nonzero fraction (an auxiliary training target; not used downstream here). Each (CT, gene) output also receives a linear skip connection from its own input gene $x_g$. 
Loss: weighted sum of $\mathrm{KL}(\bm{\pi}\,\|\,\hat{\bm{\pi}})$ on proportions and masked MSE on the expression and nonzero-fraction heads; no simplex penalty is needed because the head enforces the constraint.

\textbf{Target pair selection.} We trained on 106 biologically curated (CT, gene) pairs rather than genome-wide targets. Selection used biological heuristics and atlas marker rankings; the learnability scores were evaluated entirely post hoc:
\begin{itemize}
\item \textbf{Biological heuristics}: Prioritized immune checkpoints (CD274/PD-L1, HAVCR2/TIM-3, PDCD1/PD-1), activation/exhaustion markers (GZMB, PRF1, LAG3), tumour-microenvironment structural genes (COL1A1, FN1), and cell-type-defining genes with context-dependent expression (e.g., TBX21 in T/NK cells, CD68 in myeloid).
\item \textbf{Few stable lineage markers}: markers whose bulk variation mainly reflects abundance were deprioritized; CD19 and MS4A1 in B cells and KIT and TPSAB1 in mast cells were retained as positive controls. CT specificity alone does not imply constant or redundant within-CT expression.
\item \textbf{Avoided weak-signal targets}: Excluded ubiquitous housekeeping genes and very-low-variance genes where cell-type-specific signal is negligible.
\item \textbf{No look-ahead bias}: Pairs were selected from atlas marker rankings, not TCGA survival outcomes.
\end{itemize}
Post hoc score enrichment: selected pairs have median mean-composition plug-in $\phi^{\mathrm{CT}}_{kg}=0.41$ (IQR $0.21{-}0.68$), versus $0.12$ ($0.03{-}0.33$) for other gene--CT pairs among the 1,024 input genes. This comparator includes weakly expressed or biologically implausible combinations, so enrichment is relative to that universe, not proof that every target is accurately recoverable. The 106 pairs span 20 CTs (15 immune, 2 stromal/vascular, 3 epithelial/tumour) and 56 genes.

\subsection{TCGA NSCLC Survival Analysis}
TCGA LUAD and LUSC are combined: $n=998$ patients with OS time and event status (394 events). This predominantly resected cohort is used for exploratory prognosis, not a test of treatment response. Bulk RNA-seq is $\log_2(\mathrm{TPM}+1)$, whereas pseudobulk uses CPM, an additional source of domain shift. Predicted CTSE remains in log-expression space. Each feature and matched bulk gene is z-scored for univariate Cox (HR per SD, Wald $p$), with BH correction within family (106 CTSE, 20 proportions, 55 bulk genes). MRC1 lacks a matched bulk value because of a gene-identifier mismatch in our analysis. C-index uses 5-fold out-of-fold univariate Cox; Kaplan--Meier uses a median feature split and nominal log-rank test. Ensemble combinations use LASSO-Cox, with $\lambda$ chosen by inner 3-fold CV and evaluation by five repeats of 5-fold CV; standardisation is fitted within training folds. The bulk baseline uses the available model input genes. Adjustment and model-robustness checks are described in App.~\ref{app:controls}. Neither a significant CTSE association and a nonsignificant bulk association, nor larger point estimates, establish a significant difference between them.

\subsection{Implementation Details}
\textbf{Software:} PyTorch Lightning for model training; scikit-survival and lifelines for Cox models. \textbf{Hyperparameters:} trunk of 3 layers with 512 units (ReLU, dropout 0.1), heads with 128 hidden units (dropout 0.1); loss weight 1 for each of the two proportion heads, expression and nonzero fraction; Adam (learning rate $10^{-3}$, weight decay $5{\times}10^{-4}$), batch size 256, early stopping on held-out validation loss. \textbf{Code and data:} Training code and pseudobulk generation scripts available upon request; TCGA data publicly accessible via GDC portal (\url{https://portal.gdc.cancer.gov/}).

\end{document}